\documentclass{article}

\usepackage{arxiv}

\usepackage[utf8]{inputenc} % allow utf-8 input
\usepackage[T1]{fontenc}    % use 8-bit T1 fonts
\usepackage{hyperref}       % hyperlinks
\usepackage{url}            % simple URL typesetting
\usepackage{booktabs}       % professional-quality tables
\usepackage{amsfonts}       % blackboard math symbols
\usepackage{nicefrac}       % compact symbols for 1/2, etc.
\usepackage{microtype}      % microtypography
\usepackage{lipsum}
\usepackage{graphicx}
\usepackage{hyperref}
\usepackage{epstopdf, epsfig}
\usepackage{bm}
\usepackage{amsthm}
\usepackage{amsmath}
\usepackage[scr=esstix]{mathalpha}
\usepackage{amssymb}
\usepackage[utf8]{inputenc}
\usepackage{multirow,booktabs}
\usepackage{tabularx}
\usepackage[T1]{fontenc}
\usepackage{mathptmx}
\usepackage{multirow}
\usepackage[numbers]{natbib}
\usepackage{float}
\usepackage{eucal}
\usepackage{color}
\usepackage{subfigure}
\usepackage[ruled,linesnumbered]{algorithm2e}
\usepackage{natbib}
\usepackage{caption}
\SetKwComment{Comment}{$\triangleright$\ }{}

\title{\normalfont MD-NOMAD: Mixture density nonlinear manifold decoder for emulating stochastic differential equations and uncertainty propagation}

\author{
  Akshay Thakur \\
  Department of Aerospace and Mechanical Engineering\\
  University of Notre Dame\\
  Notre Dame, Indiana, USA.\\
  \texttt{athakur3@nd.edu} \\
   \And
 Souvik Chakraborty \\
  Department of Applied Mechanics\\
  School of Artificial Intelligence\\
  Indian Institute of Technology Delhi\\
  Hauz Khas - 110016, New Delhi, India \\
  \texttt{souvik@am.iitd.ac.in} \\
}

\begin{document}
\maketitle

\begin{abstract}
We propose a neural operator framework, termed mixture density nonlinear manifold decoder (MD-NOMAD), for emulating stochastic simulators. Our framework leverages an amalgamation of the pointwise operator learning neural architecture nonlinear manifold decoder (NOMAD) with versatile mixture density-based methods for approximating the conditional probability distributions of stochastic output functions. Furthermore, MD-NOMAD harnesses the ability of probabilistic mixture models to estimate complex probability distributions and inherits the high-dimensional scalability of NOMAD. A key feature of MD-NOMAD is its ability to perform one-shot uncertainty propagation, eliminating the need for Monte Carlo methods in statistical computations for uncertainty propagation tasks.  We conduct empirical assessments on a wide array of stochastic ordinary and partial differential equations and present the corresponding results, which highlight the performance of the proposed framework.
\end{abstract}


\section{Introduction}\label{S:1}
%Motivation

Mechanics underpins the behavior of complex physical systems and is typically modeled using differential equations (DEs). However, analytical solutions—especially for partial differential equations (PDEs)—are rarely feasible in real-world settings, leading to reliance on numerical methods such as finite difference \cite{leveque2007finite}, finite element \cite{reddy1993introduction}, and finite volume \cite{leveque2002finite} techniques. These methods can be computationally intensive, particularly when system parameters or boundary conditions are uncertain. Moreover, many physical systems involve inherent randomness, which deterministic models cannot capture. Stochastic differential equations (SDEs) address this by incorporating both deterministic and random effects. However, SDE modeling requires stochastic simulators --- computational models employing numerical techniques tailored for SDEs. Due to randomness, for fixed inputs, the outputs of stochastic simulators are random variables with unknown conditional distributions, which necessitate repeated simulations to estimate statistics or probability density functions (PDFs). A key application of SDEs is uncertainty propagation (UP) \cite{smith2013uncertainty}, which quantifies how input uncertainties affect the model outputs. Traditional methods like Monte Carlo (MC) simulations \cite{mooney1997monte}, though widely used for UP problems, require a large number of samples—even with advanced variants \cite{caflisch1998monte}—resulting in high computational costs, especially in high-dimensional settings.\par
To tackle the challenge of computational cost associated with stochastic simulators, the construction of surrogate models, which are adept at approximating the input-output relationships while maintaining computational efficiency, for these simulators has emerged as a prudent and pragmatic option. The field of surrogate modeling has garnered appreciable attention in the recent past. Within the framework of strategies inclined towards traditional methods for surrogate modeling of stochastic simulators, diverse methodologies have been investigated and refined. These include generalized additive models \cite{hastie2017generalized},  generalized linear models \cite{mccullagh2019generalized}, \cite{efromovich2010dimension}, kernel density estimators (KDE) \cite{hall2004cross,fan2018local}, Gaussian processes \cite{browne2016stochastic}, polynomial chaos expansions \cite{blatman2011adaptive}, and more sophisticated models that build upon and refine standard surrogate modeling approaches such as generalized lambda distribution-based models \cite{zhu2020replication,zhu2021emulation} or stochastic polynomial chaos expansions (SPCE) \cite{zhu2023stochastic}. However, certain limitations are associated with these methods: some are susceptible to the curse of dimensionality \cite{tsybakov2009,zhu2023stochastic,hall2004cross,browne2016stochastic}, some necessitate data to conform to specific requirements \cite{blatman2011adaptive}, while others demand numerous replications for identical input parameters \cite{zhu2020replication}. Additionally, certain methods impose restrictions on the shapes of response distributions that can be emulated \cite{hastie2017generalized,mccullagh2019generalized,zhu2021emulation}.
With the advancement of machine learning, contemporary deep learning-based methods for surrogate modeling of stochastic simulators have drawn noticeable focus. These approaches encompass various techniques, such as normalizing flow-based models \cite{lu2022learning}, generative adversarial networks (GANs) \cite{yang2020physics}, neural stochastic differential equations \cite{kidger2021neural}, denoising diffusion probabilistic models (DDPMs) \cite{ho2020denoising}, surrogate modeling through optimal transport and projection pursuit \cite{botvinick2023generative}, and gray-box methodologies like combining conditional generative moment matching networks (CGMMN) \cite{ren2016conditional} with partially known SDEs \cite{chakraborty2023deep}, or integrating neural networks (NNs) with numerical integrator-motivated loss functions to identify drift and diffusivity functions in SDEs \cite{dietrich2023learning}. Nonetheless, it's worth noting that CGMMN and GAN-based models can pose challenges during training, and normalizing flow-based models require significant computational resources \cite{botvinick2023generative}.\par

A more recent development in deep learning and surrogate modeling for DEs has been the introduction of NN architectures termed neural operators \cite{kovachki2023neural,li2020fourier,seidman2022nomad,lu2021learning,tripura2022wavelet}, which, by utilization of available data, can approximate the mapping between two continuous functions that belong to infinite-dimensional function spaces. The usage of some of the neural operator architectures is limited to specific types of meshes \cite{li2020fourier,tripura2022wavelet}, while others are mesh-independent and leverage a pointwise approximation strategy \cite{seidman2022nomad,lu2021learning}. Fourier and Wavelet neural operators \cite{li2020fourier,tripura2022wavelet} utilize discrete convolutions, which incur escalating computational costs as spatial dimensions increase. 
However, mesh-independent pointwise architectures such as NOMAD \cite{seidman2022nomad} and DeepONet \cite{lu2021learning} separate sensor measurements or parametric inputs from spatial, temporal, or spatiotemporal locations. They achieve this by feeding these two types of inputs into different sub-networks. These sub-networks are then linked through a reconstruction procedure to obtain the output. This process allows the model to produce outputs at the specified spatial, temporal, or spatiotemporal locations. Moreover, the pointwise nature of NOMAD and DeepONet also aids in scalability to higher dimensional problems. The aforementioned neural operator architectures were originally implemented as deterministic models. As such, they are more suitable as surrogate models for non-stochastic differential equations. To effectively handle stochastic simulators, we introduce a new framework --- mixture density nonlinear manifold decoder (MD-NOMAD). This pointwise framework combines mixture density modeling \cite{bishop1994mixture} with the neural operator architecture NOMAD. MD-NOMAD acts as an operator learning-based surrogate model for both stochastic ordinary and partial differential equations. More importantly, although various generative models --- such as SPCEs, GANs, and DDPMs --- can function as surrogates for stochastic simulators, they inherently operate as sample generators during the inference stage. Consequently, MC methods remain necessary to obtain the desired statistical quantities of interest. The computation of these statistics is particularly crucial for tasks such as uncertainty propagation. In contrast, the proposed MD-NOMAD framework entirely obviates the need for MC-based computations, and therefore, enables uncertainty propagation to be performed in a one-shot manner, which helps in considerable reduction in associated computational cost. Furthermore, one of MD-NOMAD's subnetworks --- the branch network --- provides a latent representation for a given parametric setting, which can be leveraged for latent space-based modeling tasks. However, the current study does not explore the utilization of the latent space. The salient features of the proposed MD-NOMAD framework are enumerated below:

\begin{itemize}
\item Due to its mesh-independent and pointwise nature, MD-NOMAD is adaptable for scaling, as a surrogate model, to high-dimensional problems solved on arbitrary meshes.
\item With an appropriate number of mixture components, MD-NOMAD is capable of modeling complex probability distributions such as multimodal, skewed, or other non-Gaussian distributions.
\item MD-NOMAD can act as a surrogate model for stochastic systems with a large number of components while employing a mixture of univariate Gaussian distributions. This is achieved by incorporating a suitable conditioning variable, which allows the decoder network of MD-NOMAD to learn the marginal distribution of individual components. This functionality of MD-NOMAD is showcased through its application to an $N$-component Lorenz-96 system.
\item Since the PDFs of the employed mixture distribution are typically accessible, the requirement for employing MC methods to estimate the statistics and PDF of the conditional output distribution during the inference stage is eliminated, and these can be computed analytically once the parameters of the mixture components are obtained from MD-NOMAD for a given input.
\item MD-NOMAD can be used to construct a surrogate model capable of performing one-shot uncertainty propagation for DEs. This capability is demonstrated on a high-dimensional elliptic stochastic partial differential equation (SPDE).
\end{itemize}

The paper is structured as follows: Section \ref{S:2} outlines the problem statement, while Section \ref{S:3} offers an overview of the necessary background for comprehending the mathematical foundations of the proposed framework, including a brief review of mixture density networks and NOMAD. Section \ref{S:4} delves into the details of the proposed framework. In Section \ref{S:5}, we present six examples to illustrate the performance of the proposed framework. Finally, Section \ref{S:6} presents the concluding remarks.
% Need of this study - start from limitation of existing studies

% What is proposed

% Organization of the pa

\section{Problem statement}\label{S:2}
Before exploring the problem statement, we initiate by introducing certain notations. Consider a fixed $N$-dimensional domain $\mathcal{D} \in \mathbb{R}^N$ bounded by $\partial \mathcal{D}$ with input function $f: \mathcal{D} \rightarrow f(\bm{x}) \in \mathbb{R}^{d_{f}}$ in the Banach space $\mathcal{F}:=C\left(\mathcal{D} ; \mathbb{R}^{d_{f}}\right)$ and output function $G: (\mathcal{D},\mathcal{F}) \rightarrow G(f)(\bm{x}) \in \mathbb{R}^{d_{G}}$ in the Banach space $\mathcal{G}:=C\left(\mathcal{D} ; \mathbb{R}^{d_{G}}\right)$, where $\bm{x}\in \mathcal{D}$. For traditional deterministic simulator, a nonlinear differential operator $\mathcal{M}$ can be defined as follows:
    \begin{equation}
    \mathcal{M}: \mathcal{F} \ni f(\bm{x}) \mapsto  G(f)(\bm{x}) \in \mathcal{G}.
    \label{equation1}
\end{equation}
In the context of pointwise mapping, considering a point $\bm{x}$ within the domain $\mathcal{D}$, the output function $G(f)(\bm{x})$ produces a real value, i.e., $d_G=1$. Additionally, in deterministic simulations, identical $\bm{x}$ and $f(\bm{x})$ inputs consistently result in the same output function values. However, with probabilistic or stochastic simulators, the uncertainty arising from the inherent characteristics of the system or insufficient knowledge thereof leads to varying output function values for a given $\bm{x}$ and $f(\bm{x})$. Therefore, to characterize the behavior of the output function for a stochastic simulator, random variable $\boldsymbol{\omega} \in \Omega$ has to be introduced. Based on this, the nonlinear differential operator for stochastic simulators can be expressed as follows:
\begin{align}
\mathcal{W}_{s}: \mathcal{F} \times \Omega &\mapsto \mathcal{G}_\omega\nonumber \\ 
(f(\bm{x}), \boldsymbol{\omega}) &\mapsto G_{\boldsymbol{\omega}}(f)(\bm{x}),
\label{equation2}
\end{align}

where $\Omega$ signifies the event space of the probability space $\{\Omega,\Sigma, \mathbb{P}\}$, encapsulating the randomness, and $G_{\boldsymbol{\omega}}(f)(\bm{x})$ is stochastic output function. Compared to deterministic simulations, stochastic simulators incur higher computational costs as numerous simulator runs have to be performed for a given set of inputs for a good characterization of PDFs for the simulator's output. Therefore, this study aims to devise a mixture density-based operator learning framework for the construction of surrogate models tailored for stochastic simulators.\par
Another objective of this study is to leverage the developed surrogate model to efficiently address the UP problem. Consider an input function \( f_{\boldsymbol{\omega}}: \mathcal{D} \times \Omega \rightarrow \mathbb{R}^{d_f} \), where \( \boldsymbol{\omega} \sim p(\boldsymbol{\omega}) \) represents parametric uncertainty. Consequently, the stochastic output function is defined as \( G(f_{\boldsymbol{\omega}})(\bm{x}, \boldsymbol{\omega}) \). Note that this notation diverges from that in Equation~\ref{equation2}, as the stochasticity is introduced through the input function; however, this scenario constitutes a specific instance of the more general formulation presented in Equation~\ref{equation2}. Within this stochastic framework, the UP problem seeks to propagate uncertainty through the nonlinear differential operator \( \mathcal{W}_s \) and to characterize the statistical properties of the resulting output \( G(f_{\boldsymbol{\omega}})(\bm{x}, \boldsymbol{\omega}) \). This involves tasks such as computing the expected value of the stochastic output,

\begin{equation}
    \mu_f = \int_{\mathcal{D}} G(f_{\boldsymbol{\omega}})(\bm{x}, \boldsymbol{\omega}) \, p(\boldsymbol{\omega}) \, d\boldsymbol{\omega},
\end{equation}

determining the variability of the stochastic output about its mean,

\begin{equation}
    \sigma_f^2 = \int_{\mathcal{D}} \left( G(f_{\boldsymbol{\omega}})(\bm{x}, \boldsymbol{\omega}) - \mu_f \right)^2 p(\boldsymbol{\omega}) \, d\boldsymbol{\omega},
\end{equation}

and constructing the PDF of the output by marginalizing over the input randomness,

\begin{equation}
    p_f(\bm{y}) = \int_{\mathcal{D}} \delta\left( \bm{y} - G(f_{\boldsymbol{\omega}})(\bm{x}, \boldsymbol{\omega}) \right) p(\boldsymbol{\omega}) \, d\boldsymbol{\omega},
\end{equation}

\noindent where \( \bm{y} \) denotes a specific realization of the stochastic output function. Solving this problem using MC methods is computationally expensive due to the extensive number of simulator or surrogate runs required to obtain adequate samples. Therefore, this work proposes a one-shot strategy employing the devised mixture density-based operator learning framework to solve the UP problem. This approach obviates the need for MC methods, thereby enabling a more tractable and efficient uncertainty propagation.

% One page on generic problem statement
\section{Preliminaries}\label{S:3}
\subsection{Mixture density networks}

Mixture Density Networks (MDNs) were introduced by Bishop \cite{bishop1994mixture} as an NN-based framework for modeling complex conditional probability distributions. Considering, $\bm{x}$ as an input to the network, MDNs model the conditional distribution $p(G_{\omega}(\bm{x}) | \mathbf{x})$, i.e, the probability distribution of model output $G_{\omega}(\bm{x})$  given input $\bm{x}$, using a mixture of $\mathit{m}$ probability distributions. Mathematically, the mixture-based approximation of conditional distribution using NNs can be expressed as 

\begin{equation}
p(G_{\omega}(\bm{x}) | \bm{x}) = \sum_{i=1}^{m} \pi^i(\bm{x};\boldsymbol{\theta}) \cdot \phi(G_{\omega}(\bm{x}); \rho^i(\bm{x};\boldsymbol{\theta}))
\label{equation3}
\end{equation}

\noindent Here, $\pi^i(\bm{x};\boldsymbol{\theta})$ is the mixing coefficient of the $i$-th component and $\phi(G_{\omega}(\bm{x}); \rho^i(\bm{x};\boldsymbol{\theta}))$ represents the PDF of the $i$-th component, where the parameters of PDF and mixing coefficients are in turn parameterized by NN parameters $\boldsymbol{\theta}$ and also depend on the input $\bm{x}$. Furthermore, any parametric family of probability distributions, for example, Gaussian, Laplacian, beta, exponential, or others, can be used as the mixture components, and the choice is contingent upon the characteristics of the problem being modeled. 

% The mixture components can follow any parametric family of distributions, such as Gaussian, exponential, or others, depending on the problem context and requirements. However, a common choice is to use Gaussian distributions due to their simplicity and tractability. In this case, $f(y; \boldsymbol{\theta}_k(\mathbf{x}))$ represents a Gaussian distribution with mean $\boldsymbol{\mu}_k(\mathbf{x})$ and covariance $\boldsymbol{\Sigma}_k(\mathbf{x})$, parameterized by $\boldsymbol{\theta}_k(\mathbf{x}) = (\boldsymbol{\mu}_k(\mathbf{x}), \boldsymbol{\Sigma}_k(\mathbf{x}))$.

During training, the parameters of MDN are optimized by utilizing the input-output training data pairs to predict the mixture components parameters ($\pi^i(\bm{x};\boldsymbol{\theta})$ and $\rho^i(\bm{x};\boldsymbol{\theta})$). This is typically achieved by minimizing the negative log-likelihood loss function. Furthermore, during inference, the MDN can generate samples from the conditional distribution $p(G_{\omega}(\bm{x}) | \bm{x})$ for a given input $\bm{x}$ by sampling from the mixture-based distribution or can provide PDFs or requisite statistics using analytical computations based on the mixture component parameters. 

\subsection{NOMAD}

A widely adopted approach for constructing operator learning frameworks, as proposed by Lanthaler et al. \cite{lanthaler2022error}, involves composing three approximation maps. By employing this composition, the operator $\mathcal{M}$ defined in Equation \ref{equation1} can be approximated as follows:
\begin{equation}    
    \mathcal{M}: \mathcal{R}\circ \mathcal{A} \circ \mathcal{E}.
\end{equation}
Here, $\mathcal{E}$ represents an encoding map that transforms a continuous function to its evaluation at $\mathscr{p}$ locations. The encoder is succeeded by the application of the approximation map, represented by $\mathcal{A}: \mathbb{R}^{\mathscr{p}} \rightarrow \mathbb{R}^{\mathscr{h}}$, which maps pointwise function evaluations to a latent space with dimension $\mathscr{h}$. Generally, $\mathcal{A}$ is an NN. The composition of the encoding and approximation maps denoted as $\beta(f) = \mathcal{A} \circ \mathcal{E}(f)$, is referred to as the branch network. This is followed by the reconstruction or decoding map, denoted as $\mathcal{R}:\mathbb{R}^{\mathscr{h}} \rightarrow \mathbb{R}^{d_G}$, which maps the finite-dimensional latent feature representation to the output function. In a DeepONet, the reconstruction map is a dot product between a set of functions, $\tau_k \in C(\mathcal{D}; \mathbb{R}^{d_{G}})$, parameterized by an NN (also known as the trunk network) and $\beta(u)(x_k)$, where $d_g=1$ and $k=1,2,\ldots \mathscr{h}$. This can be expressed as 
\begin{equation}
\mathcal{R}(\boldsymbol{\beta}) = \sum_{k=1}^\mathscr{h} \tau_k \beta_k.
\end{equation}
However, the utility of linear reconstruction maps is constrained by the Kolmogorov n-width \cite{seidman2022nomad,pinkus2012n}. To address this limitation, Seidman et al. \cite{seidman2022nomad} introduced a nonlinear reconstruction/decoding map that is parameterized by an NN. This map can be expressed as follows:
\begin{align}
\mathcal{R}(\boldsymbol{\beta},\bm{x}) =  \mathcal{K}(\boldsymbol{\beta}, \bm{x}),
\label{equation4}
\end{align}
where, $\mathcal{K}: \mathbb{R}^{\mathscr{h}} \times \mathcal{D} \rightarrow \mathbb{R}^{d_G}$ is a deep neural network. Ultimately, the fusion of the nonlinear decoder with the branch network constitutes the NOMAD architecture.

\section{Proposed framework}\label{S:4}
\begin{figure}[htbp!]
    \centering
    \includegraphics[width=1.0\textwidth]{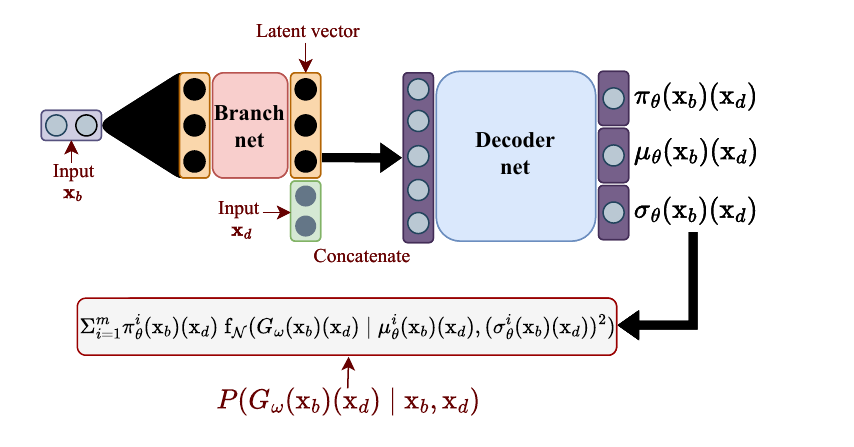}
    \caption{Schematic diagram depicting the architecture of MD-NOMAD. Here, \( \mathbf{x_b} \) serves as the input to the branch network, which is an FCN producing a finite-dimensional latent feature representation as its output. The sensor values or locations \( \mathbf{x_d} \) (also referred to as augmenting input) are augmented to the latent vector obtained from the branch network and provided as an input to the decoder network, which is also an FCN. The output of the decoder network comprises the parameters of the $\mathscr{m}$-component mixture of a probability distribution, with the probability distribution chosen from a suitable parametric family. In this article, we use a mixture of univariate Gaussian distribution. Subsequently, the MD-NOMAD model is trained by minimizing the sum of the negative logarithm of the probability \( P(G_{\omega}(\mathbf{x_b})(\mathbf{x_d})) \) over \( \mathscr{n} \) training samples.}
    \label{fig:1}
\end{figure}
As discussed in Section \ref{S:3}, while a deterministic NOMAD can effectively learn a mapping from input function evaluations to output function values, it cannot directly model the mapping for a stochastic operator, which typically necessitates the estimation of the PDF of the output function. Hence, to address this, we combine the pointwise operator learning NOMAD framework with the mixture density-based formalism to arrive at the mixture density NOMAD (MD-NOMAD) architecture. This MD-NOMAD framework, by design, enables the estimation of the conditional probability distribution for a stochastic output function \(G_{\omega}(f)(\bm{x})\) associated with the input function \(f(\bm{x})\) given a sensor value \(\bm{x}\). Although a variety of parametric distributions can be used for the mixture components, the choice largely depends on factors such as the nature of the response distribution being modeled. Some distribution families are better suited for capturing skewness or heavy tails, while others may offer advantages in terms of training stability, interpretability, or simplicity. In this study, we limit our focus to the univariate Gaussian distribution, chosen for its balance of simplicity, robustness during training, and interpretability. Thus, within this framework, an analog to \(\rho_i(\bm{x};\boldsymbol{\theta})\) in Equation \ref{equation3} is represented as \(\rho_{\theta}^i(\mathbf{x_b})(\mathbf{x_d})\), denoting a Gaussian distribution with mean \(\mu_{\boldsymbol{\theta}}^i(\mathbf{x_b})(\mathbf{x_d})\) and standard deviation \(\sigma_{\theta}^i(\mathbf{x_b})(\mathbf{x_d})\) (or variance \(\sigma_{\boldsymbol{\theta}}^i(\mathbf{x_b})(\mathbf{x_d})^2\)) for the \(i\)-th mixture component, where the mean, standard deviation/variance, along with the mixing coefficient, are parameterized by a NOMAD with parameters \(\boldsymbol{\theta}\).  In addition, to ensure that the standard deviation/variance is non-negative, softplus activation function is applied to the NOMAD predicted component parameters for standard deviation/variance. Furthermore, \(\mathbf{x_b}\) and \(\mathbf{x_d}\) represent the inputs to the branch and decoder network, respectively. A schematic of the MD-NOMAD's architecture is illustrated in Figure \ref{fig:1}. 

The optimization of the trainable parameters of MD-NOMAD, given \(\mathscr{n}\) input-output data pairs, which enables it to predict the mixture parameters, is accomplished through the minimization of an appropriate loss function \(\mathcal{L}(\mathcal{G}_{\omega}, \mathcal{G}_{\omega})\). In addition, in surrogate models for stochastic simulators, which inherently involve probabilistic elements, one typically employs probabilistic loss functions like Kullback-Leibler divergence, maximum mean discrepancy, or negative log-likelihood. However, in the context of MD-NOMAD, our focus lies in determining the parameters of the component distribution that maximize the likelihood of observed or provided data given these parameters. Consequently, in the current scenario, the minimized loss function is the negative log-likelihood, expressed as:

\begin{equation}
\mathcal{L}(G_{\omega}^{1:\mathscr{n}}(\mathbf{x_b})(\mathbf{x_d})| \mathbf{x_b},\mathbf{x_d}, \boldsymbol{\theta}) = - \sum_{k=1}^{\mathscr{n}}  \log \left( \sum_{i=1}^{\mathscr{m}} \pi_{\theta}^i(\mathbf{x_b})(\mathbf{x_d}) \cdot f_{\mathcal{N}}(G_{\omega}(\mathbf{x_b})(\mathbf{x_d}); \rho_{\theta}^i(\mathbf{x_b})(\mathbf{x_d})) \right),
\end{equation}

\noindent where $f_{\mathcal{N}}$ is a Gaussian probability density function.  It is key to note that MD-NOMAD inherits scalability characteristics from NOMAD (or DeepONet) architectures owing to their point-wise approach, which is particularly beneficial for high-dimensional problems. Additionally, the complexity of our architecture, specifically, the final layer involving parameters of the mixture densities, is contingent upon the number of components employed. This, in turn, is influenced by the complexity of the probability distribution at sensor locations. Moreover, once the model can predict the parameters of component distributions for given inputs upon successful optimization, the computations for statistics and PDFs for the model-predicted conditional output distribution can be done analytically.
In addition, throughout this study, we adopt a replication-based experimental design to generate training and validation data. Initially, we create a distinct set of input parameters by sampling from predetermined parameter ranges. The size of this set is represented by $\mathscr{N}$, while $\mathscr{R}$ denotes the number of replications for each input parameter. Together, this collection of unique input parameters and their replications constitutes the experimental design. Subsequently, for each point within the experimental design, a corresponding target output is generated utilizing an appropriate stochastic simulator tailored to the specific problem under investigation.

% \begin{algorithm}[t]
% \caption{Mini-batch gradient descent \label{alg:one}
% \textbf{Input:} Input and response dataset
% \textbf{Output:} Learned parameter $\boldsymbol{\theta}$
% \end{algorithm}

\section{Results and discussion}
\label{S:5}
In this section, we present an empirical evaluation of the performance of the proposed model across various stochastic differential equations, encompassing both linear and nonlinear formulations, alongside an analytical benchmark problem. Furthermore, given that the proposed framework involves parameter prediction for a mixture of Gaussian distributions, the predicted PDFs and statistics by the proposed model for the target variables in all examples presented in this article, given a specific input set, rely on analytical computation rather than MC-based sampling methodologies. For example, the predicted PDF is obtained by computing the sum of Gaussian PDFs (equal to the number of components used) weighed by the corresponding mixture weights and parameterized by the corresponding mean and standard deviation/variance values predicted from MDN-NOMAD. Mathematically, the computation of a PDF for a particular set of $\mu$, $\sigma$, and $\pi$ values predicted by an $m$-component MD-NOMAD can be expressed as follows
\begin{equation}
f(x) = \sum_{i=1}^{m} \pi_i \cdot\frac{1}{\sigma_i\sqrt{2\pi}} e^{-\frac{(x-\mu_i)^2}{2\sigma_i^2}}.
\end{equation}
In addition, we showcase the utility of our model for one-shot uncertainty propagation within a setting based on a 2-dimensional SPDE. The primary objective of this study is to efficiently obtain the PDF of a system’s response, thereby enabling a comprehensive characterization of its stochastic behavior and informed decision-making. In contrast to DDPM, the proposed MD-NOMAD framework offers a faster analytical approach for estimating the response PDF for any given input conditioning. Additional details regarding this comparative computational efficiency are provided in Appendix ~\ref{app}. A similar rationale underlies the relatively slower performance of SPCE in deriving the response density. Therefore, we benchmark the performance of MD-NOMAD against a KCDE-based model for the problems under investigation. Although KCDE, akin to SPCE, is affected by the curse of dimensionality, it possesses the advantage of not requiring sampling prior to density estimation. The conditional response distribution in KCDE is estimated as follows:

\begin{equation}
\hat{f}_{Y|\bm{X}}(y \mid \bm{x}) =
\frac{
\sum\limits_{i=1}^{n} K_{\bm{X}}\left( \bm{x}, \bm{x}_i,h_x \right) K_{Y}\left(y,y_i,h_y \right)
}{
\sum\limits_{j=1}^{n} K_{\bm{X}}\left( \bm{x}, \bm{x}_j,h_x \right)
},
\end{equation}

where \( K_{\scalebox{1.5}{$\cdot$}} \)'s are the kernel functions and \( h_{\scalebox{1.5}{$\cdot$}} \)'s their corresponding bandwidth parameters, selected via grid search. To compare the performance of the KCDE-based model and the MD-NOMAD surrogate, we employ an error metric based on the Wasserstein distance of the second order. This metric is defined as:

\begin{equation}
    \mathcal{E}_{W} = \mathbb{E}_{\bm{X}}\left[l^{W}_2(Y_{\bm{X}},\hat{Y}_{\bm{X}})\right],
\end{equation}

where \( Y_{\bm{x}} \) represents the simulator response, \( \hat{Y}_{\bm{x}} \) denotes the response of the surrogate model, and \( l^{W}_2 \) is the squared 2-Wasserstein distance, expressed as:

\begin{equation}
l^{W}_2(Y_1, Y_2) = \| Q_1 - Q_2 \|^2_2 
= \int_0^1 (Q_1(v) - Q_2(v))^2 \, dv,
\end{equation}

with \( Q_1 \) and \( Q_2 \) representing the quantile functions of the random variables \( Y_1 \) and \( Y_2 \), respectively. We also employ KL-divergence ($\mathcal{E}_{KL}$) \cite{wang2009divergence} as an error metric. Both of these error metrics are calculated on an unseen test set for each of the example problems, utilizing both the MD-NOMAD and KCDE-based model as emulators, and the results are detailed in Table \ref{tab:model-comparison}. Additionally, the hyperparameters associated with MD-NOMAD, including the number of mixture components, number of layers, and learning rate, are determined through a grid search for each problem.
%and its adaptability for ensemble-based uncertainty quantification techniques.
\subsection{Stochastic Van der Pol equation}
The Van der Pol equation, characterized by non-conservative dynamics and nonlinear damping, is utilized across diverse domains such as biology, electronics, physics, and economics. However, in this example, we utilize the stochastic variant of this equation, which can be mathematically expressed as follows
\begin{equation}
\begin{aligned}
dX_t &= Y_t \, dt, \\
dY_t &= \mu(1 - X_t^2)Y_t \, dt + \lambda^2 X_t \, dW_t,\\
\text{where}\; X_0 &= -3,\; Y_0 = 0,\;\mu = 1,\; \lambda \sim \mathcal{U}(0.4,0.8),\; t \in [0,20].
\end{aligned}
\end{equation}
Here, $dW_t$ is used to represent the Wiener noise. We are only interested in the construction of a surrogate for the temporal evolution of $X_t$, and thus, we ignore $Y_t$. One of the interesting aspects of choosing the mentioned parametric sampling range for $\lambda$ is that this choice allows for bifurcation in our quantity of interest (QOI), i.e., $X_t$, as it evolves with time, and therefore, the corresponding PDF undergoes a transition from being unimodal to being bimodal. %Additionally, the PDF exhibits sharper peaks at initial times, which gradually diffuse over the modeled time-span. 
The stochastic Van der Pol equation is solved using order 1.0 strong Stochastic Runge-Kutta algorithm \cite{rossler2010runge}, and the python-based package sdeint is leveraged. For this example, the experimental design is contingent on stochastic noise strength $\lambda$, i.e, we sample $\mathscr{N}$ values for $\lambda$ from the decided sampling distribution, replicate them $\mathscr{R}$ times, and solve the SDE for each value to generate the required dataset.\par
We use the MD-NOMAD to learn the mapping from $[\lambda,t]$ to $[\pi(\lambda)(t),\,\mu(\lambda)(t),\,\sigma(\lambda)(t)]$. The mapping is eventually employed to obtain the probability density function $f_{X|\Lambda,T}(x)(\lambda,t)$. We use a small two-layered fully connected network (FCN) for the branch network and a comparatively larger 4-layered FCN for the decoder network.  In addition, the stochastic noise strength is employed as an input to the branch net whereas $t$ is used as the augmenting input to the decoder network. Due to our choice for the experimental design, the complete dataset consists of $\mathscr{N}\times \mathscr{R}$ solution trajectories. Furthermore, the full dataset is randomly split into training and validation datasets with a $9:1$ size ratio. The Adam optimizer is employed for the mini-batch-based training with a suitable learning rate and schedule, while the number of training epochs and mixture components is set to $300$ and $15$ respectively. Also, the number of layers in the branch and decoder network along with the optimizer and the train-validation split is kept the same, unless stated otherwise, for different problems assessed in this study. In addition, the number of realizations and replications used for the current example are  $\mathscr{N} = 50$ and  $\mathscr{R} = 40$, respectively.\par
\begin{figure}[ht!]
\centering
\subfigure[]{\label{subfig:lab21}\includegraphics[width=1.0\textwidth]{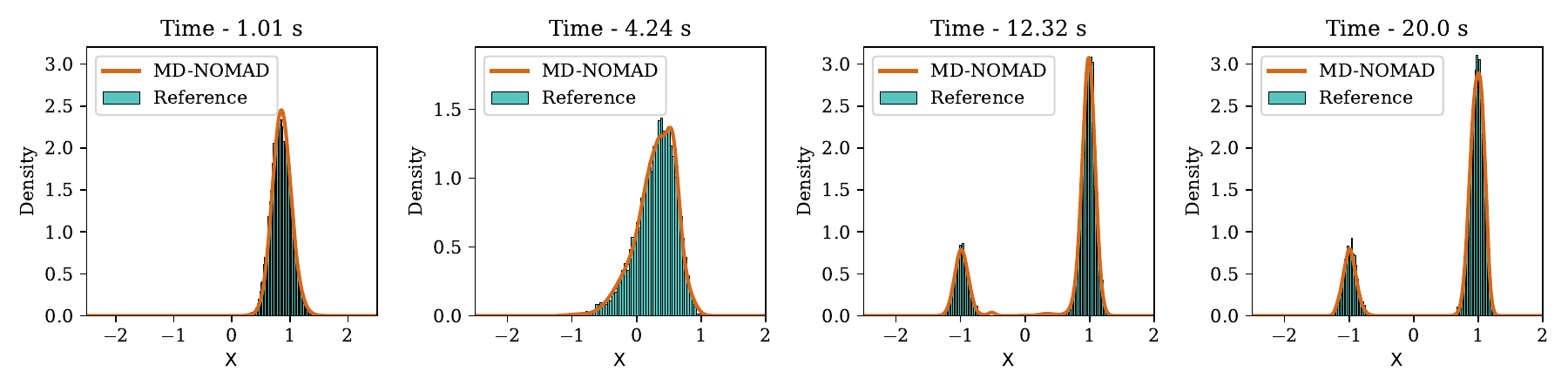}}
\subfigure[]{\label{subfig:lab22}\includegraphics[width=1.0\textwidth]{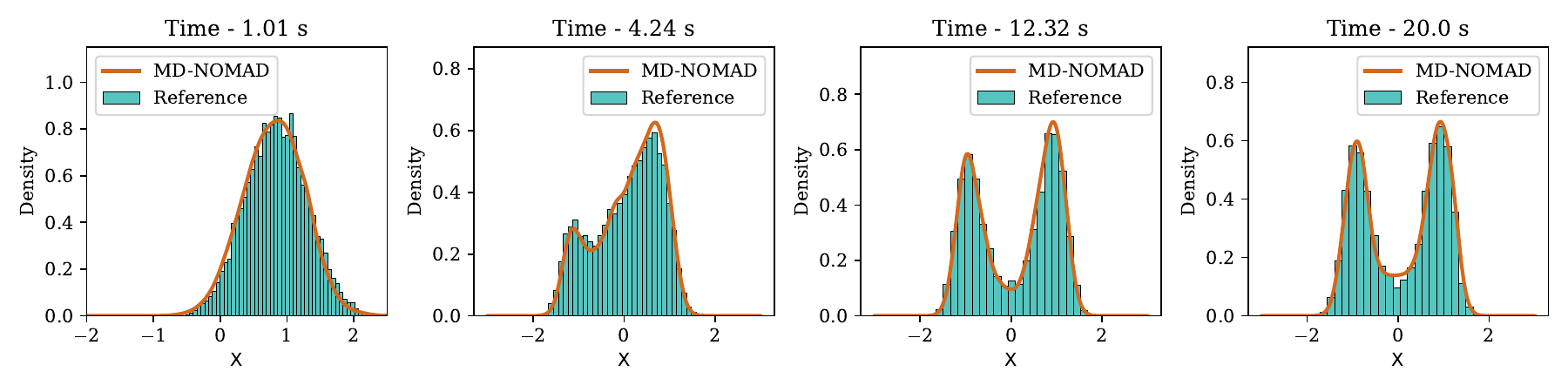}}
\caption{Comparison between reference distribution and MD-NOMAD predicted PDF at times $t = {1.01,\, 4.24,\, 12.32,\,\text{and } 20}$ s for  \textbf{(a)} $\lambda = 0.45$, \textbf{(b)} $\lambda = 0,75$ for the spatial location component of stochastic Van der Pol equation.}
\label{fig:2}
\end{figure}
\begin{figure}[ht!]
\centering
\subfigure[]{\label{subfig:lab31}\includegraphics[width=0.48\textwidth]{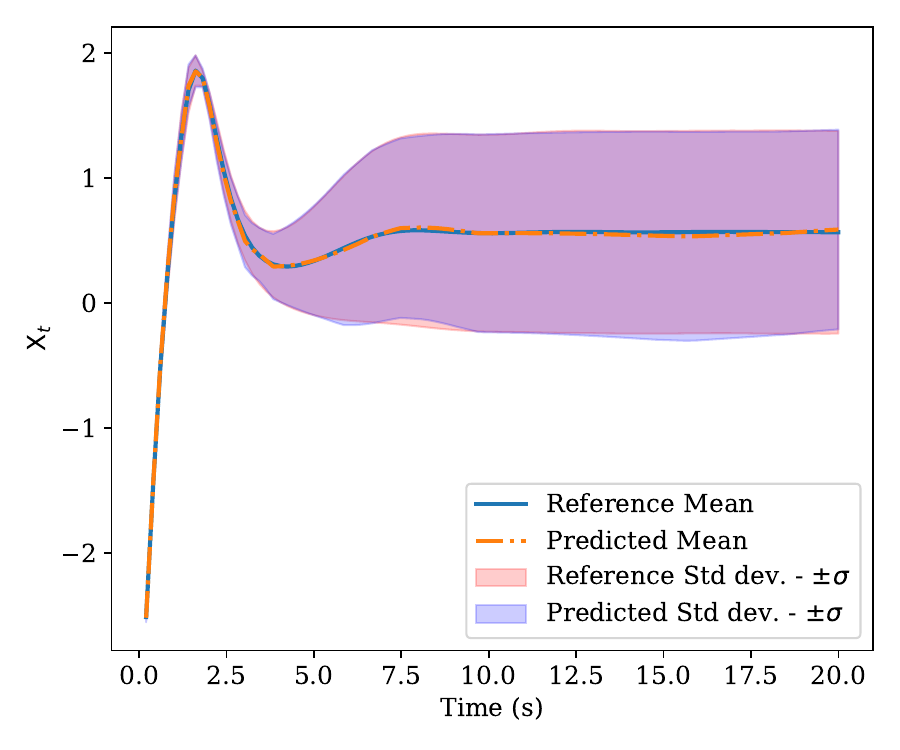}}
\subfigure[]{\label{subfig:lab32}\includegraphics[width=0.48\textwidth]{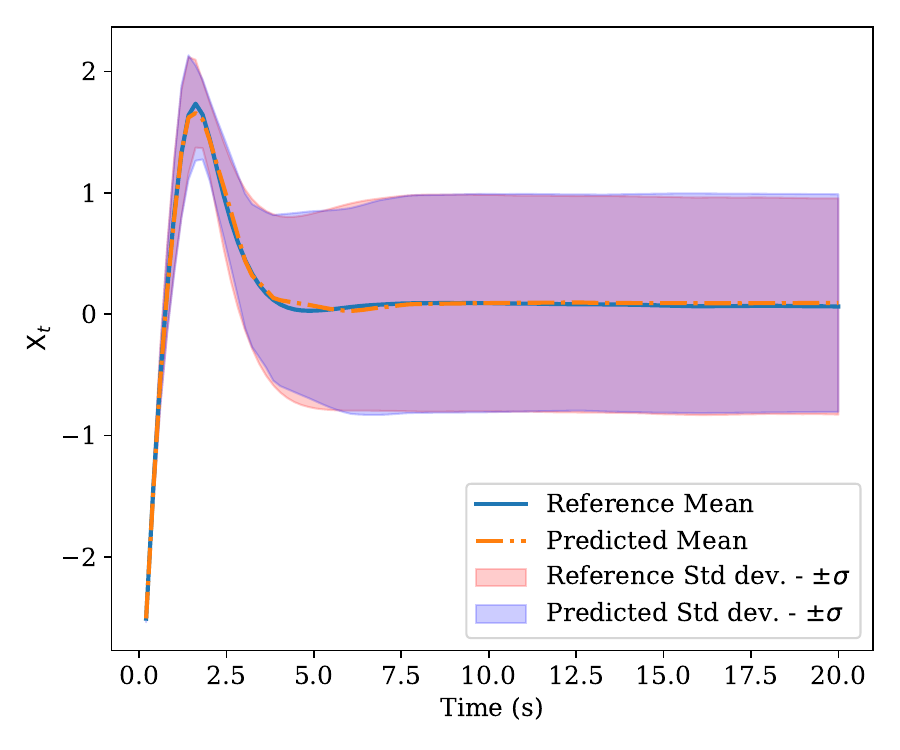}}
\caption{Comparison between reference statistics and MD-NOMAD predicted statistics till $t=20$ s for  \textbf{(a)} $\lambda = 0.45$, \textbf{(b)} $\lambda = 0,75$ for the spatial location component of stochastic Van der Pol equation.}
\label{fig:3}
\end{figure}

Finally, to assess the performance of the trained MD-NOMAD, we perform a comparison between the reference distribution and statistics with the model-predicted PDF and the statistics. We carry out this comparison on two randomly chosen values for stochastic noise strength $\lambda = 0.45$ and $\lambda = 0.75$, which are not present in the training or the validation dataset. Specifically, two types of pictorial comparisons are performed --- the first one being the comparison between the predicted PDF and the reference distribution for $X_t$ at a few different temporal locations through the solution time span to ascertain how well the proposed model performs in capturing multiple moments of the desired distribution, and the second one is the comparison between the reference statistics, i.e., mean and standard deviation, and the predicted statistics for the entire solution time-span. Both the reference distribution and corresponding statistics are obtained from the empirical distribution of $10^4$ replications. The samples are generated using the same stochastic simulator utilized for producing the training data, both for the current case and for all other example problems presented in this study. We again emphasize that as the output of MD-NOMAD is the parameters for a mixture of Gaussian distribution, the predicted PDF and statistics are obtained in an analytic way obviating the need for a large number of MC simulation (MCS) runs. The results for comparison between the predicted PDF and the reference distribution at different temporal locations are presented in Figure \ref{fig:2}, while the comparison between the temporal evolution of predicted and reference statistics is presented in Figure \ref{fig:3}. It can be observed from Figure \ref{fig:2} that MD-NOMAD can predict PDFs for $X_t$ with reasonable accuracy even after the bifurcation, i.e., where the PDF transitions from unimodal to bimodal. 

\subsection{N-component stochastic Lorenz-96}

The stochastic Lorenz-96 system represents a nonlinear set of stochastic ordinary differential equations, offering a simplified model to capture atmospheric dynamics, particularly the intricate multi-scale interactions. We consider the $N$-component stochastic Lorenz-96 system, expressed as:
\begin{equation}
    \begin{aligned}
     dX^i_t &= [(X^{i+1}_t - X^{i-2}_t)X^{i-1}_t - X^i_t + F]dt + \lambda \cdot dW^i_t,
    \end{aligned}
\end{equation}

\noindent where $i \in \{1,2,\ldots,N\}$ denotes cyclic indices. For this illustration, we aim to construct a surrogate for all $N$-components of the state variable $X_t$. The trajectories for the Lorenz-96 system are generated utilizing the Euler-Maruyama method with a time step $dt = 0.04$ over the period $t = [0, 4] s$. In addition, the experimental design is based on the diffusion coefficient $\lambda$, which is sampled from a uniform distribution $\mathcal{U}(0.15,0.35)$ with $\mathscr{N} = 50$ realizations and $\mathscr{R}=30$ replications. Despite the introduction of a general $N$-component system, our investigation confines itself to two scenarios: $N=10$ and $N=50$. It's important to recognize that this system also experiences bifurcation, and the PDFs for different components of $X_t$ transition from densities with sharper peaks to diffused ones as time progresses.\par
\begin{figure}[ht!]
\centering
\subfigure[]{\label{subfig:lab41}\includegraphics[width=1.0\textwidth]{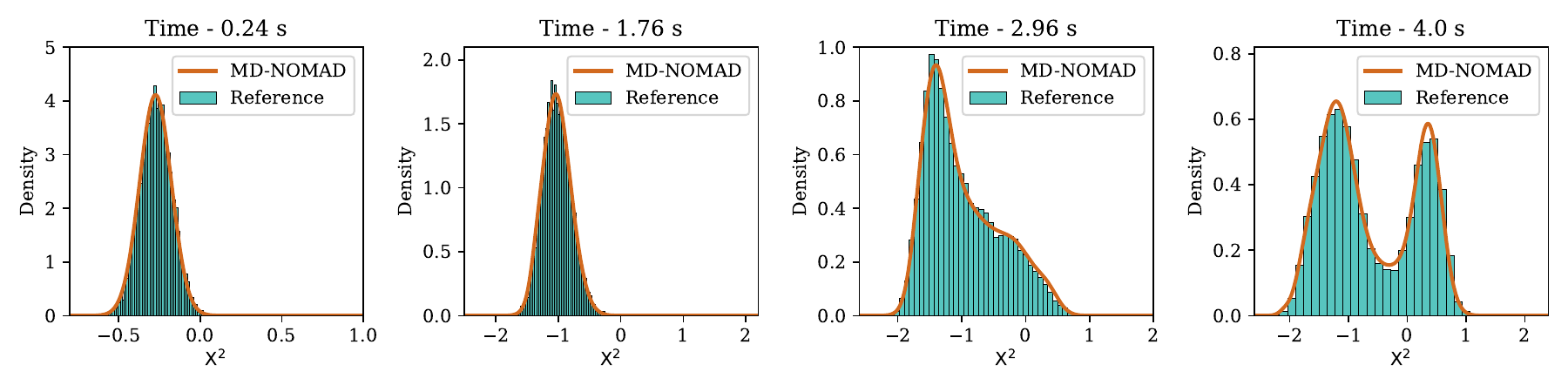}}
\subfigure[]{\label{subfig:lab42}\includegraphics[width=1.0\textwidth]{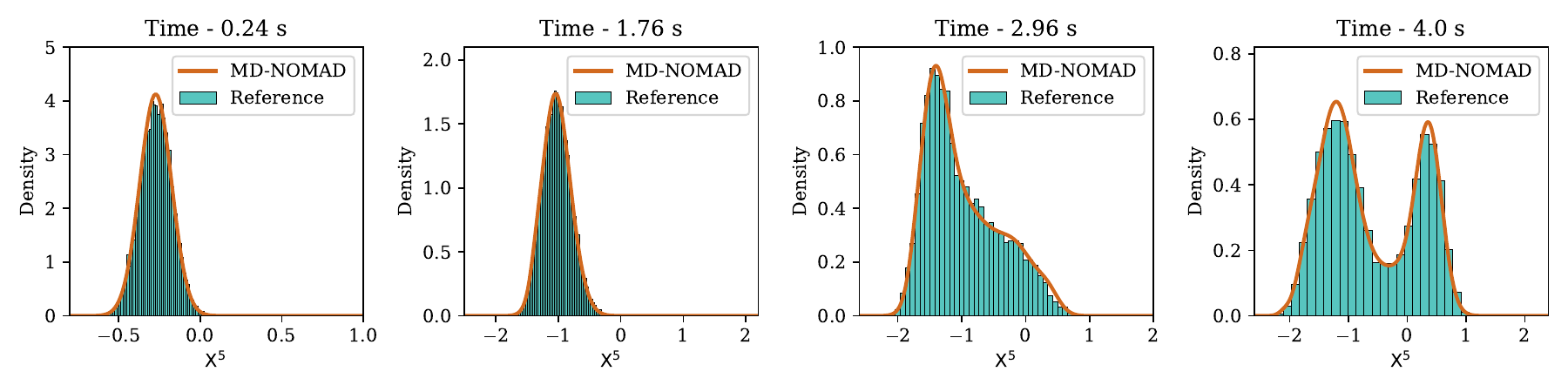}}
\subfigure[]{\label{subfig:lab43}\includegraphics[width=1.0\textwidth]{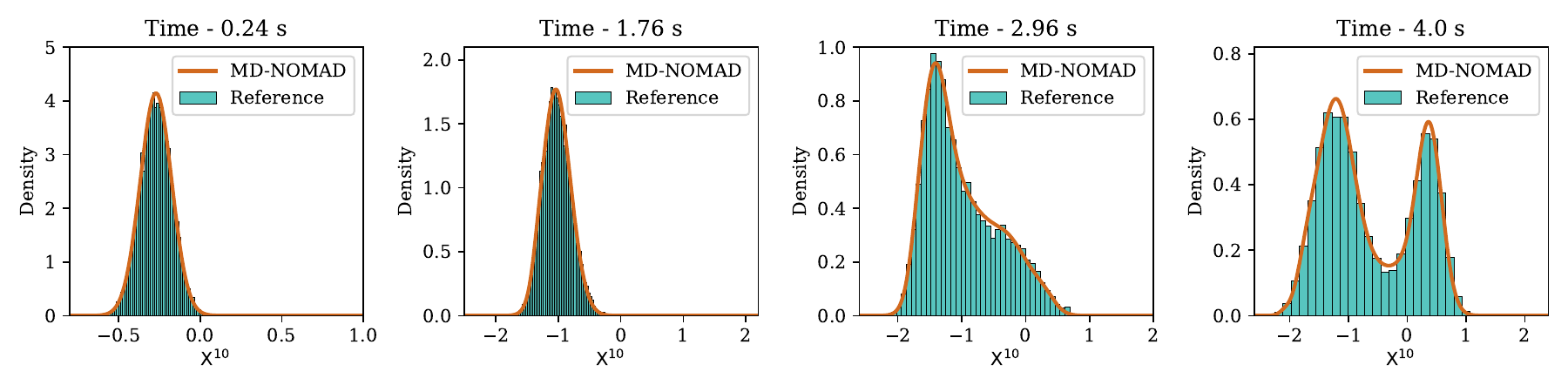}}
\caption{Comparison between reference distribution and MD-NOMAD predicted PDF at times $t = {0.22,\, 1.76,\, 2.96,\,\text{and } 4.0}$ s corresponding to $\lambda=0.22$ for the \textbf{(a)} second, \textbf{(b)} fifth, and  \textbf{(c)} tenth solution component of the $10$-component stochastic Lorenz equations.}
\label{fig:4}
\end{figure}
\begin{figure}[ht!]
\centering
\subfigure[]{\label{subfig:lab51}\includegraphics[width=1.0\textwidth]{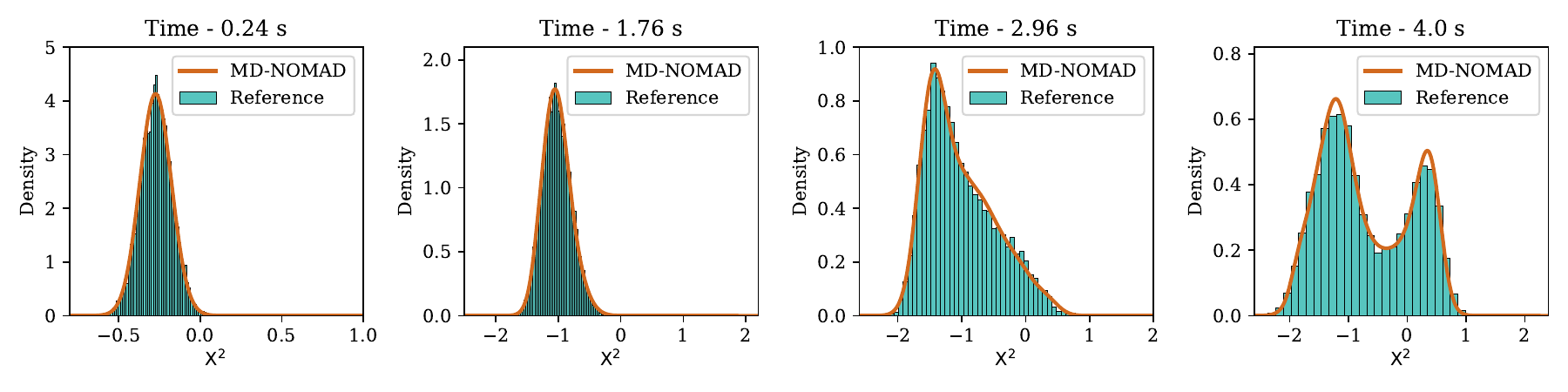}}
\subfigure[]{\label{subfig:lab52}\includegraphics[width=1.0\textwidth]{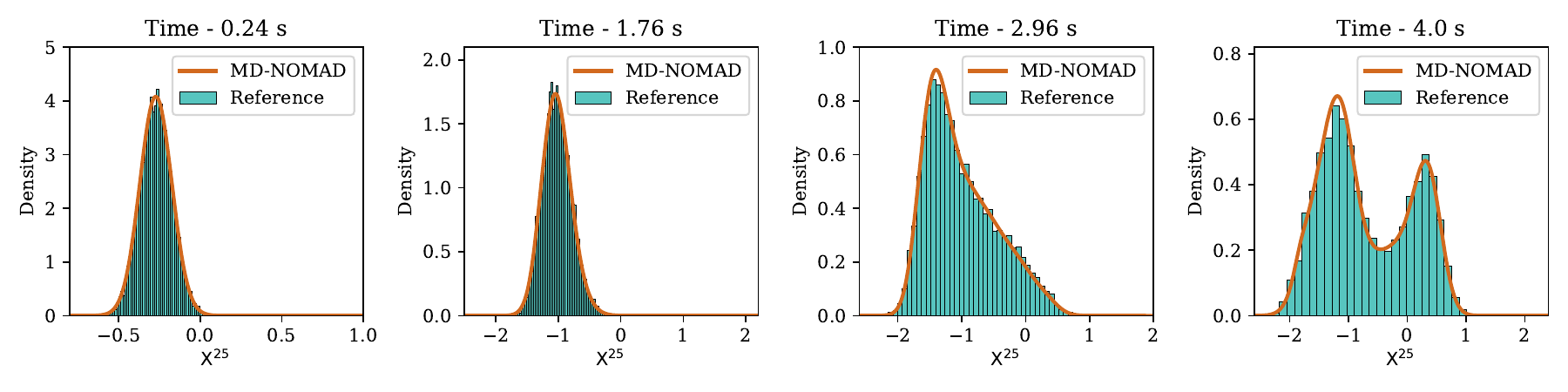}}
\subfigure[]{\label{subfig:lab53}\includegraphics[width=1.0\textwidth]{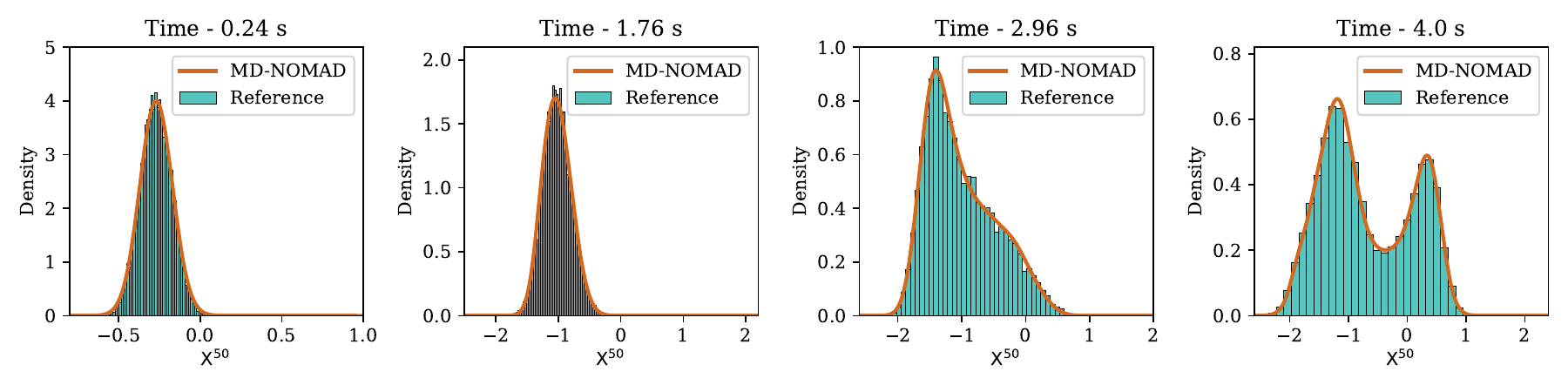}}
\caption{Comparison between reference distribution and MD-NOMAD predicted PDF at times $t = {0.22,\, 1.76,\, 2.96,\,\text{and } 4.0}$ s corresponding to $\lambda=0.22$ for the \textbf{(a)} second, \textbf{(b)} twenty fifth, and  \textbf{(c)} fiftieth solution component of the $50$-component stochastic Lorenz equations.}
\label{fig:5}
\end{figure}
\begin{figure}[ht!]
    \centering
    \includegraphics[width=1.0\textwidth]{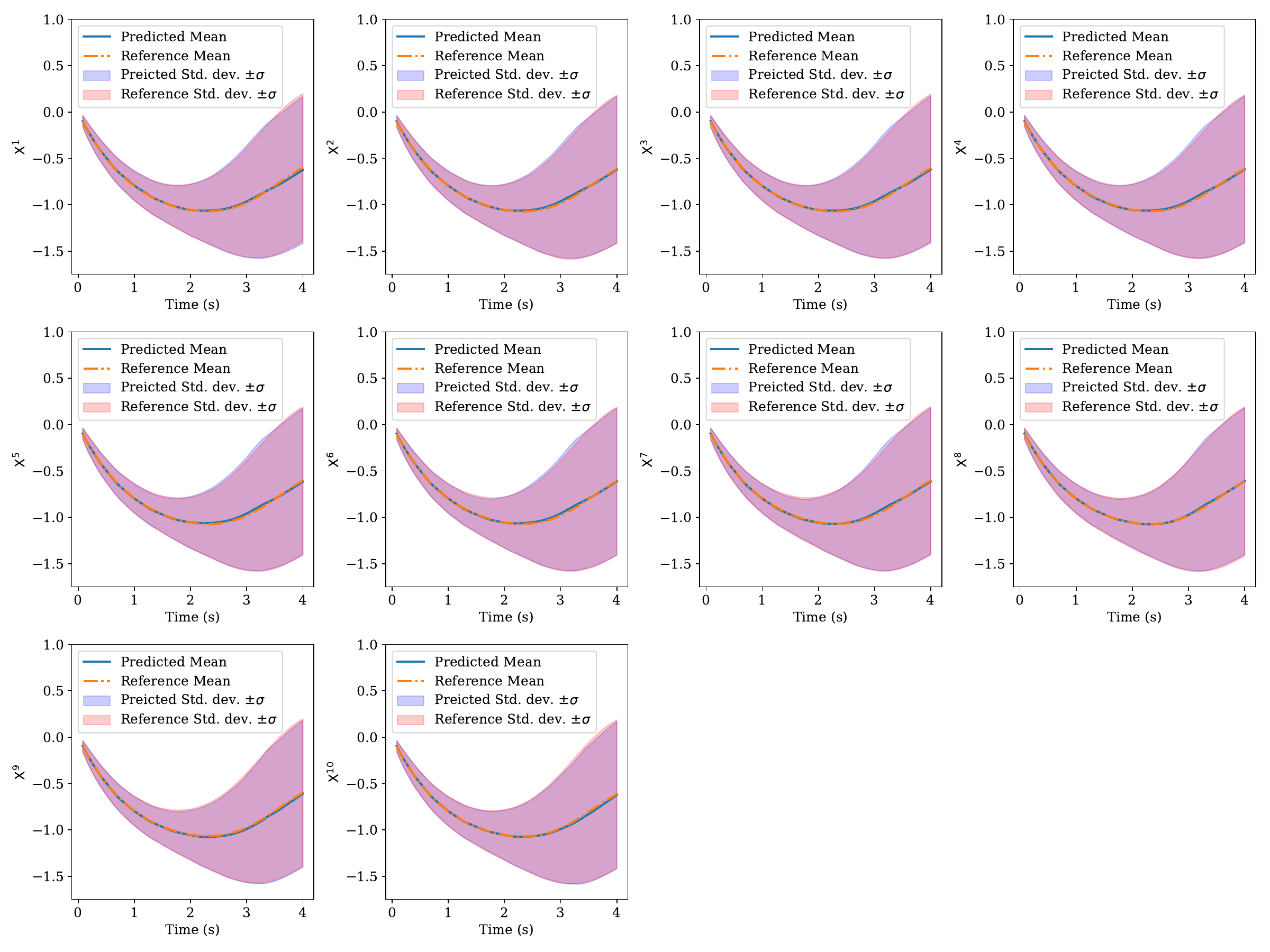}
    \caption{Comparison between reference statistics and MD-NOMAD predicted statistics till $t=4$ s for all ten components ($X^i$) of the $10$-component stochastic Lorenz equations corresponding to the test case with $\lambda=0.22$.}
    \label{fig:6}
\end{figure}
\begin{figure}[ht!]
    \centering
    \includegraphics[width=0.9\textwidth]{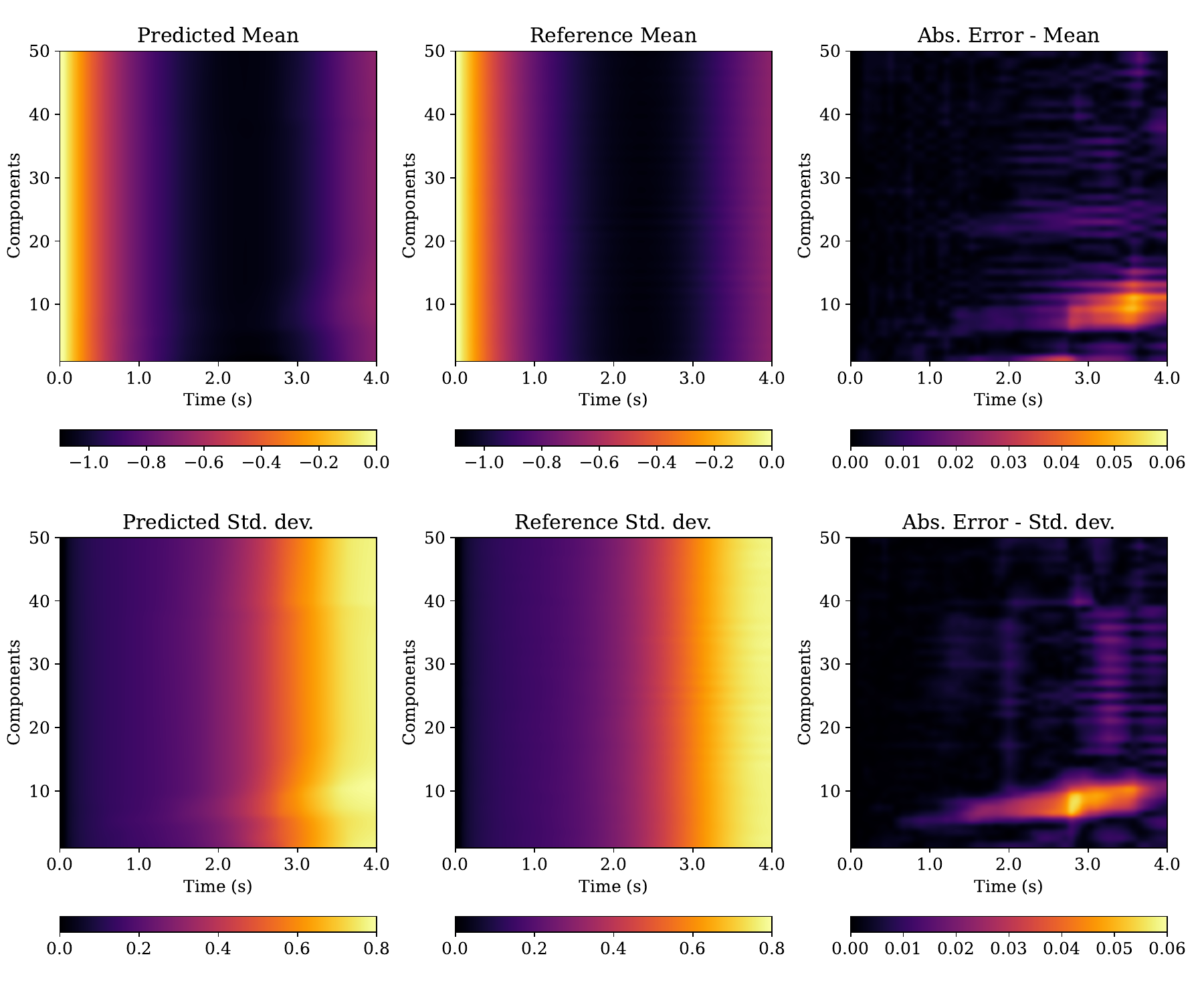}
    \caption{Comparison between reference statistics and MD-NOMAD predicted statistics till $t=4$ s for all ten components ($X^i$) of the $50$-component stochastic Lorenz equations corresponding to the test case with $\lambda=0.22$. For compendiousness, the mean and standard deviation values, for a time span of 4 s and all the components (in an ascending component number order), are vertically stacked to obtain a 2-d image.}
    \label{fig:7}
\end{figure}

The MD-NOMAD, for this system of equations, is used to construct a surrogate for the following mapping:  $[\lambda,t,x_d]$ $\mapsto$ $[\pi(\lambda)(t,x_c),\,\mu(\lambda)(t,x_c),\,\sigma(\lambda)(t,x_c)]$.  We use $\lambda$ as the branch network's input, while $(t, x_d)$ are concatenated to latent branch network outputs to form the inputs for the decoder network. As we employ a pointwise NN architecture in this study, we require conditioning variables for different components $X^i$ of the $N$-component Lorenz equation. In this example, $x_d \in [1,2,\ldots,N]$ acts as a conditioning variable for enabling MD-NOMAD to construct the required surrogate map for a particular component. It is important to understand that $x_d$ has to be scaled suitably before supplying it to the decoder network. We scale $x_d$ by dividing the size of Lorenz's system, i.e., $N$. Furthermore, during the preprocessing of the training and validation dataset, the system dynamics trajectories are temporally subsampled by a factor of $2$ to mimic a setting with sparser data availability. The number of components used for the Gaussian mixture and the total number of training epochs is chosen to be $15$ and $250$ for the $50-$component system, whereas, for the $10-$component system these are set to $10$ and $250$. For conciseness, the results containing the predictions for the PDFs at different times, which can be found in Figure \ref{fig:4} and Figure \ref{fig:5}, are presented for some selected components. Furthermore, the results containing the statistics are presented in Figure \ref{fig:6} and Figure \ref{fig:7}. The results correspond to a case with stochastic noise strength $\lambda = 0.22$, which is not a part of the training and validation dataset.  Furthermore, the reference
distribution and corresponding statistics are obtained from the empirical distribution of $10^4$
replications.


\subsection{Bimodal analytical benchmark}
In this example, we extend a complex analytical case introduced by Zhu and Sudret \cite{zhu2023stochastic} by incorporating additional parametric dependence on mixture weights. The target distribution is defined as:
\begin{equation}
    f_{Y|X,\Lambda}(y)(X=x,\Lambda = \lambda)=\lambda~\mathit{\phi}(4\sin^{2}(\pi\cdot x)+4x-2)+(1-\lambda)~\mathit{\phi}(4\sin^{2}(\pi\cdot x)-4x+2).
    \label{eq:53}
\end{equation}

\noindent Here, $\mathit{\phi}$ denotes a normal PDF with a nonlinear mean function with respect to $x$ and a standard deviation of $0.8$. The parameter $\lambda$ parameterizes the mixture coefficients, and $X$ is an input vector of $x$-values consisting of $100$ uniformly spaced points between $0$ and $1$. The target PDF is a mixture of two Gaussian PDFs. The experimental design is based on the mixture coefficient parameter $\lambda$, which is sampled from a uniform distribution $\mathcal{U}(0.4,0.7)$ with $\mathscr{N} = 70$ realizations and $\mathscr{R}=30$ replications. The number of training epochs and the number of mixture component is equal to $300$ and $10$ respectively. The objective is to construct the following mapping with MD-NOMAD:  $[\lambda,x]$ $\mapsto$ $[\pi(\lambda)(x),\,\mu(\lambda)(x),\,\sigma(\lambda)(x)]$. The complementary pair $(\lambda, 1-\lambda)$ is used as an input to the branch network, while $x$ is appended to the latent output of the branch network to create an input for the decoder network.

\begin{figure}[ht!]
    \centering
    \includegraphics[width=1.0\textwidth]{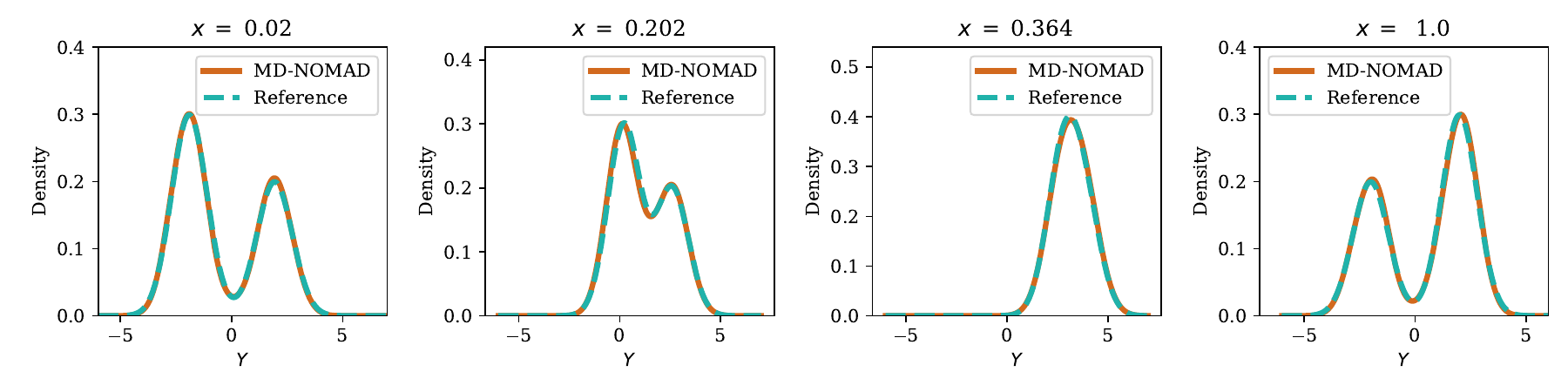}
    \caption{Comparison between reference distribution and MD-NOMAD predicted PDF at $x = (0.02,\, 0.202,\, 0.364,\, 1.0)$ corresponding to $\lambda=0.6$ for the analytical bimodal benchmark.}
    \label{fig:8}
\end{figure}

\begin{figure}[ht!]
    \centering
    \includegraphics[width=0.53\textwidth]{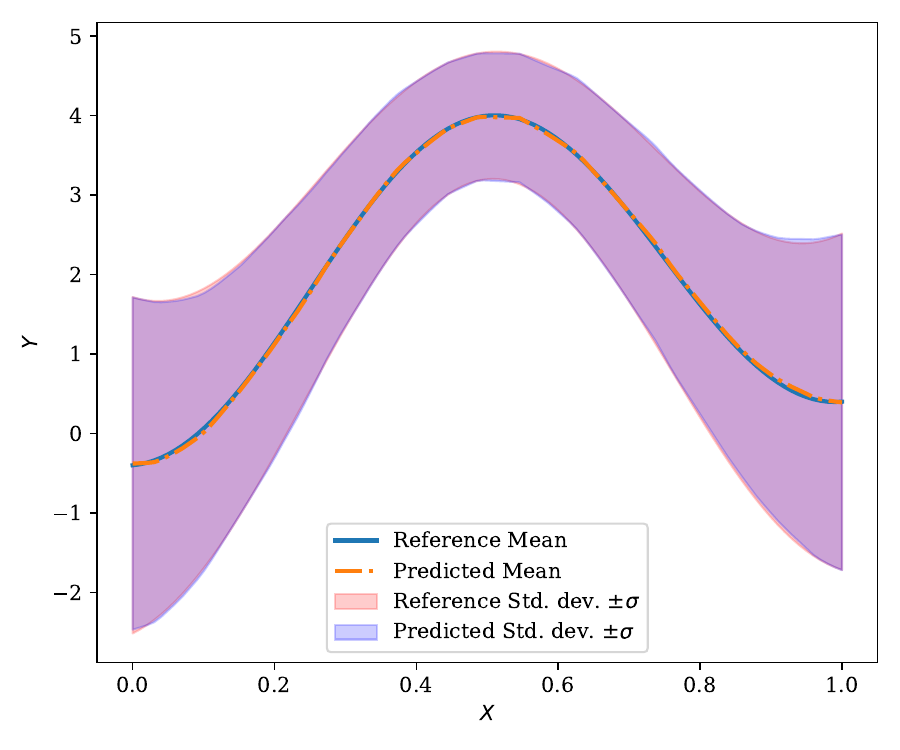}
    \caption{Comparison between reference statistics and MD-NOMAD predicted statistics from $x=0$ to $x=1$ s for analytical bimodal benchmark corresponding to the test case with $\lambda=0.6$.}
    \label{fig:9}
\end{figure}

For the evaluation of our framework in a testing scenario, we set $\lambda$ to be equal to $0.6$. Interestingly, the resulting conditional distribution transitions from being bimodal for $x \lesssim 0.2$ to unimodal for $0.2 \lesssim x \lesssim 0.8$ and again back to bimodal for $x\gtrsim0.8$ \cite{zhu2023stochastic}. Additionally, the reference PDFs used for comparison with the predicted PDFs are obtained analytically using Equation \ref{eq:53}. The pictorial illustration of the performance of MD-NOMAD is provided in Figure \ref{fig:8} and Figure \ref{fig:9}.

\subsection{2-dimensional SPDE: One-shot uncertainty propagation}
Consider the following 2-dimensional elliptic PDE:
\begin{equation}
-\nabla\cdot(\alpha(x,y) \nabla u(x,y))=0, \quad \text{where } (x, y) \in \mathcal{D},
\end{equation}
with $\alpha(x,y)$ representing the diffusion coefficient. This equation operates within a unit square domain, where $\mathcal{D} \in [0,1]^{2}$, and it is adapted from Tripathy and Bilionis \cite{tripathy2018deep}. Furthermore, the equation outlined above is subject to the following boundary conditions:
\begin{equation*}
\begin{aligned}
u &= 0, & \text{for } x &= 1, \\
u &= 1, & \text{for } x &= 0, \\
\frac{\partial u}{\partial n} &= 0, & \text{for } y &= 0 \text{ and } y = 1.
\end{aligned}
\end{equation*}
The stochasticity, which accounts for the associated uncertainty, is introduced through the diffusion coefficient $\alpha(x,y)$. The diffusion coefficient is modeled as a log-normal random field, expressed as:
\begin{equation}
\log \alpha(x,y) \sim \operatorname{GP}\left(\alpha(x,y) \mid m(x,y), \mathcal{K}\left((x, y),\left(x^{\prime}, y^{\prime}\right)\right)\right),
\end{equation}
where the mean function is $m(x,y) = 0$ and the covariance function $\mathcal{K}\left((x, y),\left(x^{\prime}, y^{\prime}\right)\right)$ is modeled using the Matérn 3/2 kernel which can be expressed as:
\begin{equation}
\mathcal{K}\left((x, y),\left(x^{\prime}, y^{\prime}\right)\right) = \sigma^2 \left(1 + \sqrt{3}\,\frac{r\left((x, y),\left(x^{\prime}, y^{\prime}\right)\right)}{l_x}\right)\exp\left(-\sqrt{3}\,\frac{r\left((x, y),\left(x^{\prime}, y^{\prime}\right)\right)}{l_y}\right),
\end{equation}
where $r\left((x, y),\left(x^{\prime}, y^{\prime}\right)\right) = \sqrt{(x-x^{\prime})^{2} + (y-y^{\prime})^{2}}$ represents the Euclidean distance between points $(x, y)$ and $(x^{\prime}, y^{\prime})$. Also, $l_x$ and $l_y$  are the lengthscale parameters in $x$ and $y$ directions, respectively. The experimental design is based on the lengthscale pair $(l_x, l_y)$ denoted as $\ell_{xy}$. The number of realizations denoted as $\mathscr{N}$, for $\ell_{xy}$, i.e., the number of unique pairs of $\ell_{xy}$, is set to $50$. These unique pairs are sampled from a beta distribution, $\text{Beta}(\alpha=1.2, \beta=4)$, to introduce a sampling bias towards smaller lengthscales. Additionally, $l_x$ and $l_y$ are constrained within the range $[0.05, 0.9]$, where the lower limit is set such that it is larger than the grid size. Furthermore, these pairs are replicated $50$ times, i.e., $\mathscr{R} = 50$. After the generation of the experimental design, discretization is performed on the problem domain to obtain a $32 \times 32$ grid. Subsequently, samples are generated for the diffusion field, $\alpha \in \mathbb{R}^{32 \times 32}$, on the grid corresponding to each point in the experimental design. This is succeeded by solving the PDE using the finite volume method to obtain the numerical solution $\mathbf{\hat{u}} \in \mathbb{R}^{32 \times 32}$.\par
\begin{figure}[ht!]
\centering
\subfigure[]{\label{subfig:lab101}\includegraphics[width=1.0\textwidth]{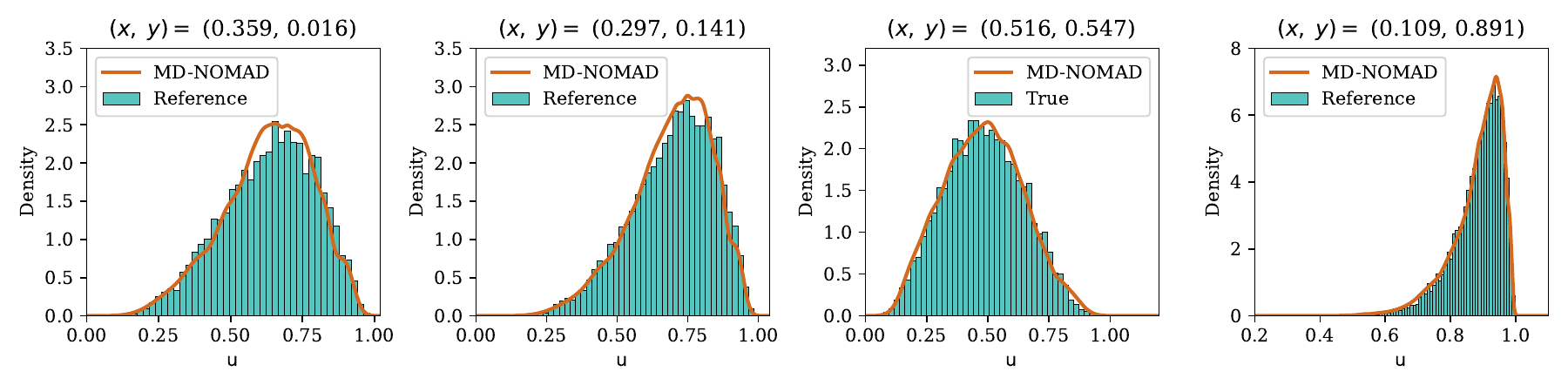}}
\subfigure[]{\label{subfig:lab102}\includegraphics[width=1.0\textwidth]{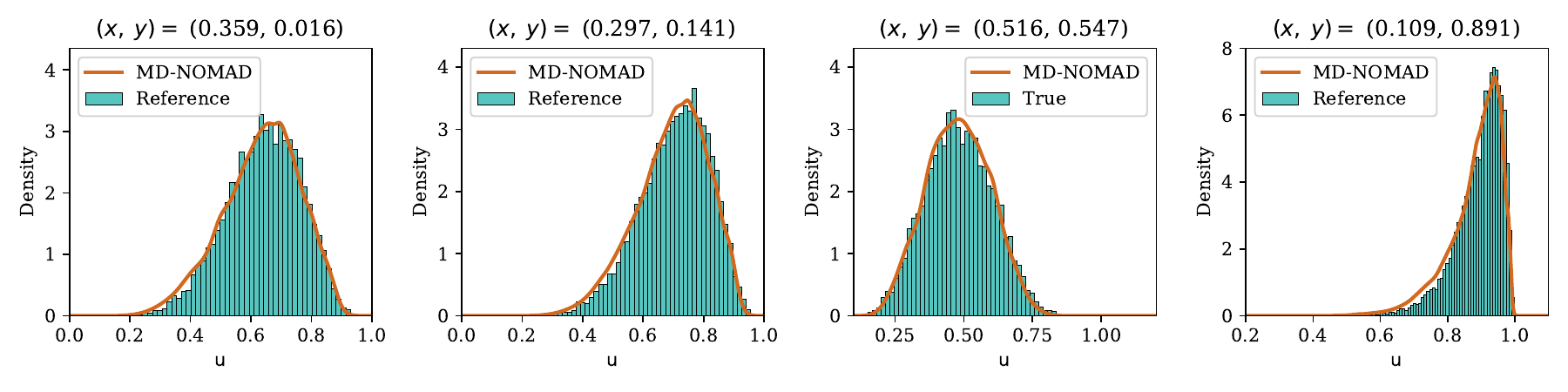}}
\caption{Comparison between reference distribution and MD-NOMAD predicted PDF at different spatial locations corresponding for \textbf{(a)} $\ell_{xy} = (0.3,0.15)$ and \textbf{(b)} $\ell_{xy} = (0.06,0.12)$ for uncertainty propagation problem associated with 2-d elliptic SPDE.}
\label{fig:10}
\end{figure}
\begin{figure}[htbp!]
\centering
\subfigure[]{\label{subfig:lab111}\includegraphics[width=0.69\textwidth]{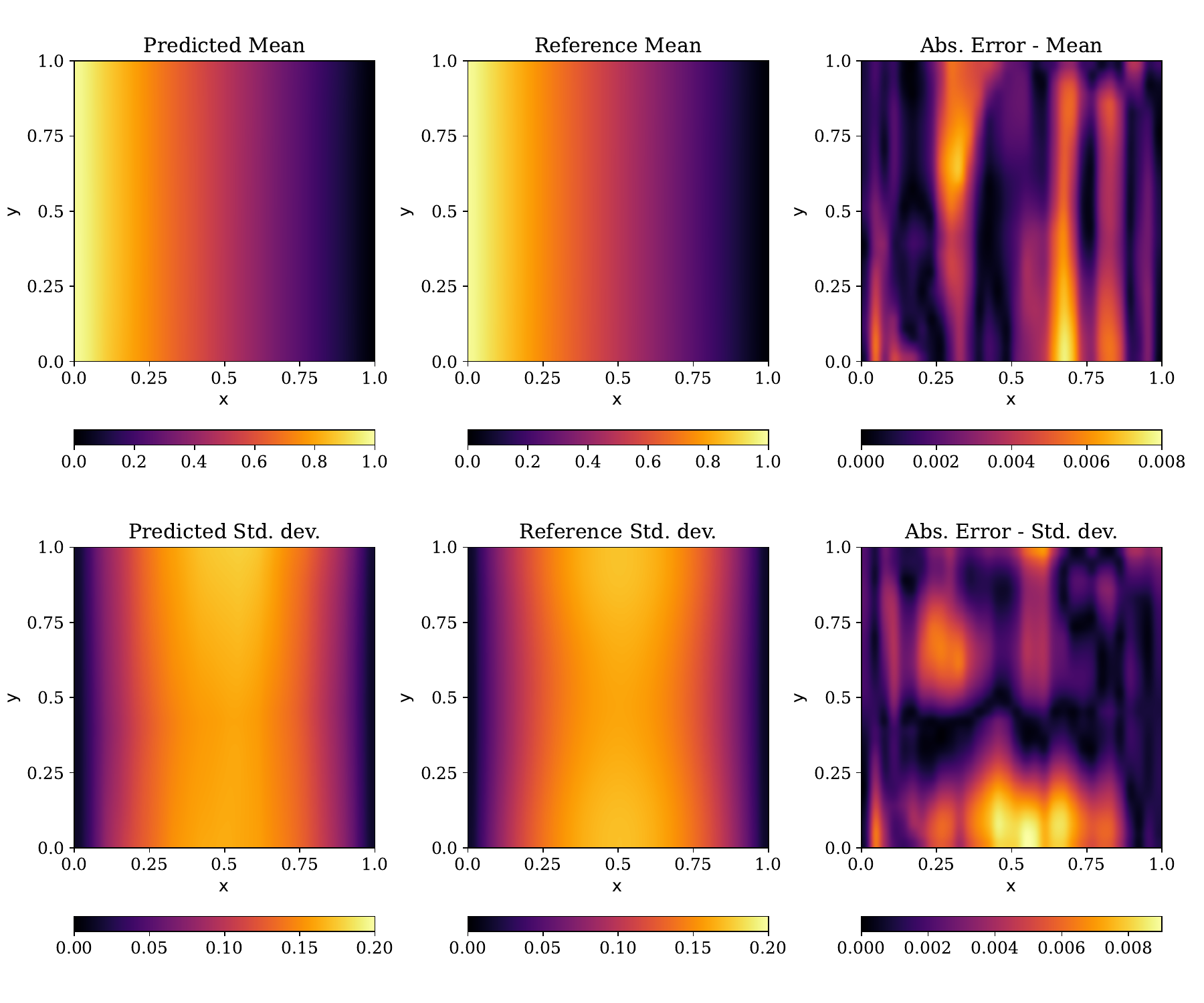}}
\vspace{-1em}
\subfigure[]{\label{subfig:lab112}\includegraphics[width=0.69\textwidth]{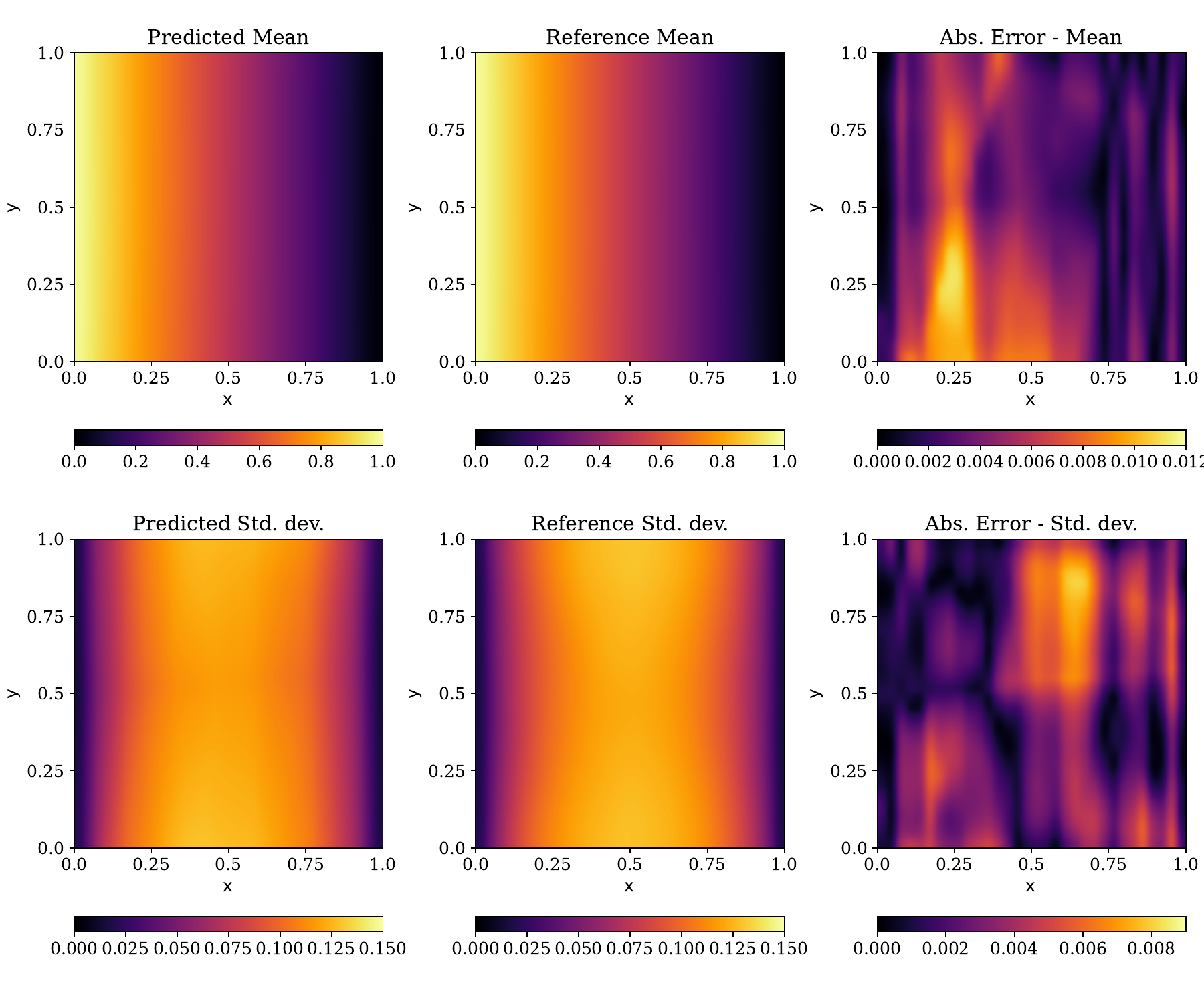}}

\caption{Comparison between reference statistics and MD-NOMAD predicted statistics for entire spatial domain corresponding for \textbf{(a)} $\ell_{xy} = (0.3,0.15)$ and \textbf{(b)} $\ell_{xy} = (0.06,0.12)$ for uncertainty propagation problem associated with 2-d elliptic SPDE.}
\label{fig:11}
\end{figure}

Our aim with this problem is to directly learn the map from lengthscale pair, $\ell_{xy}$, to the parameters of the distribution, i.e.,$[\pi(\ell_{xy})(x,y),\,\mu(\ell_{xy})(x,y),\,\sigma(\ell_{xy})(x,y)]$, of the uncertain solution for a given location $(x,y)$, thereby propagating the uncertainty with one forward evaluation of the MD-NOMAD. The number of training epochs and the number of mixture component is equal to $300$ and $50$ respectively. Furthermore, $\ell_{xy}$ is used as an input to the branch network and $(x,y)$ is used as augmenting input for the decoder network. Also, the train and validation data split for this example is $80:20$. We assess the performance of the trained MD-NOMAD on two different test lengthscale pairs --- $\ell_{xy} = (0.3,0.15)$ and $\ell_{xy} = (0.06,0.12)$.  The results are presented in Figure \ref{fig:10} and Figure \ref{fig:11}. The reference
distribution and corresponding statistics are obtained from the empirical distribution of $10^4$
replications. Furthermore, we also compare the performance of MD-NOMAD with deterministic NOMAD (DNOMAD) using the relative $\ell_1$-error between the predicted statistics. The corresponding results are included in Table \ref{tab:1}. Moreover, for DNOMAD, the statistics are estimated by using MCS and $10^4$ trials had to be performed, whereas the statistics and even the PDFs presented in Figure \ref{fig:10} were obtained with analytical computations. 

\begin{table}[htbp!]
\caption{Relative $\ell_1$-error for mean and standard deviation values of the uncertain SPDE solution corresponding to different test lengthscale pairs.}
\label{tab:1}
\centering
\small
\renewcommand{\arraystretch}{1.25}
\begin{tabular}{l c c c c}
\hline\hline
\multirow{2}{*}{\textbf{Model}} &
\multicolumn{2}{c}{$\boldsymbol{\ell_{xy} = \left(0.3, 0.15\right)}$} &
\multicolumn{2}{c}{$\boldsymbol{\ell_{xy} = \left(0.06, 0.12\right)}$} \\
\cline{2-5}
& Mean & Std. dev. & Mean & Std. dev. \\
\hline
MD-NOMAD & 0.0062 & 0.0251 & 0.0090 & 0.0455 \\
DNOMAD & 0.0108 & 0.1071 & 0.0141 & 0.1955 \\
\hline\hline
\end{tabular}
\normalsize
\end{table}

\subsubsection{Computational cost comparison between the MC method, KDE, and MD-NOMAD}
In this section, we evaluate and compare the computational cost associated with the Monte Carlo (MC) method, Kernel Density Estimation (KDE), and MD-NOMAD for estimating probability density functions (PDFs) and statistical quantities, with computational time serving as the primary metric. In general, statistical computations using the MC method require a large number of samples, typically on the order of \(10^4\). Consequently, to estimate statistics at a given spatial location \((x, y)\) for the two-dimensional SPDE, substantial sample generation is necessary. Similarly, KDE also relies on a large sample size to produce accurate and smooth PDF estimates. In contrast, MD-NOMAD eliminates the need for sample generation altogether, thereby achieving a significant reduction in computational overhead. This advantage is clearly reflected in the wall-clock times reported for each method in the estimation of PDFs and statistical measures, as presented in Table~\ref{tab:comp_time}.
\subsubsection{Demonstration of resolution invariance in MD-NOMAD}
\begin{figure}[ht!]
    \centering
    \includegraphics[width=1.0\textwidth]{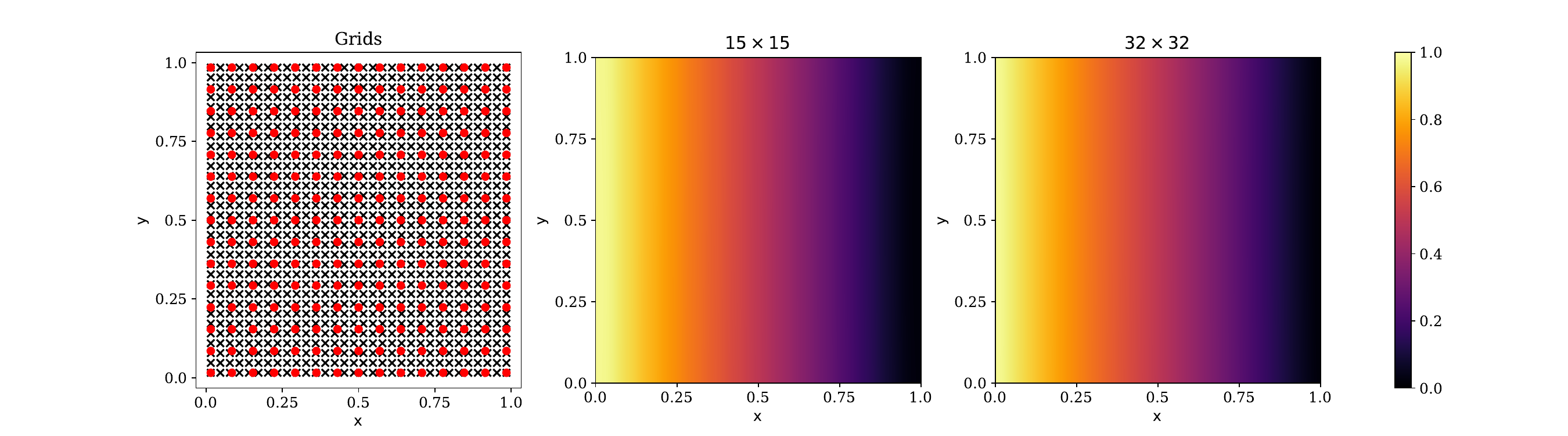}
    \caption{Illustration of resolution invariance in MD-NOMAD. (Left) Comparison of the training and testing grids: black crosses denote the original $32 \times32$ training grid points, and red circles indicate the $15 \times15$ testing grid points with different spatial locations. (Middle) Predicted mean field on the $15 \times15$ testing grid. (Right) Predicted mean field on the original $32 \times32$ training grid. Mean fields correspond to the length scale $\ell_{xy} = (0.3,0.15)$.}
    \label{fig:11a}
\end{figure}

In this section, we demonstrate the resolution invariance of MD-NOMAD by evaluating its performance on a different grid resolution from that used during training. Specifically, the model was originally trained on a 
$32 \times32$ grid. We now test it on an UP problem defined over a $15 \times15$ grid, with spatial locations that do not overlap with the training grid. Figure~\ref{fig:11a} presents the predicted mean fields for both the training and testing grids corresponding to the length scale $\ell_{xy} = (0.3,0.15)$, along with the grids.
\begin{table}[ht!]
\caption{Computational time comparison between MC-method, KDE, and MD-NOMAD for PDF and statistical computations for the 2-dimensional SPDE.}
\centering
\begin{tabular}{l c c}
\hline\hline
\textbf{Method} & \textbf{PDF (s)} & \textbf{Statistics (s)} \\
\hline
MD-NOMAD & 0.037 & 0.398 \\
\hline
KDE and MC          & 301.406 & 301.847 \\
\hline\hline
\end{tabular}

\label{tab:comp_time}
\end{table}

\subsection{1-dimensional stochastic heat equation}
\begin{figure}[ht!]
\centering
\subfigure[]{\label{subfig:lab121}\includegraphics[width=1.0\textwidth]{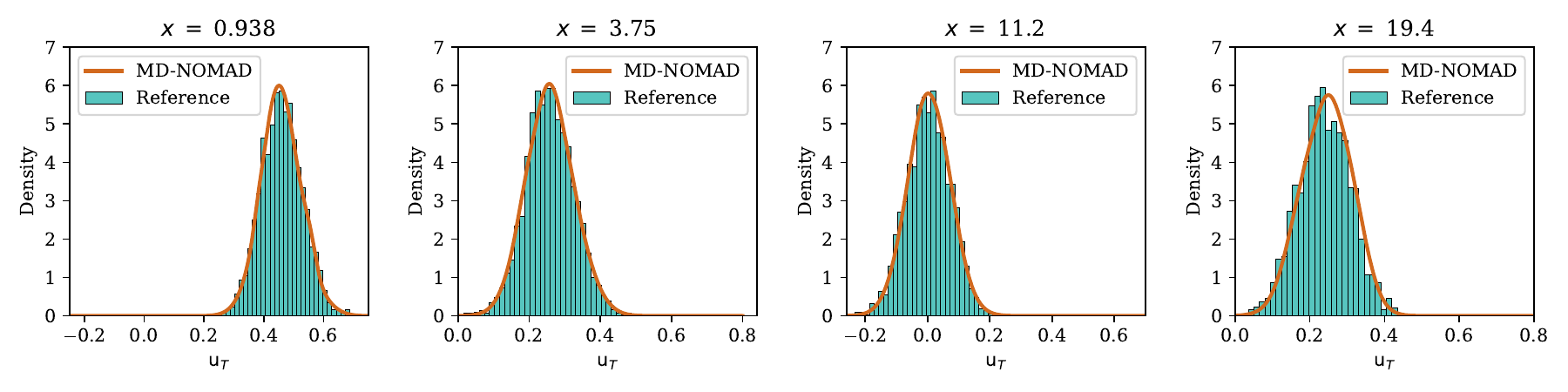}}
\subfigure[]{\label{subfig:lab122}\includegraphics[width=1.0\textwidth]{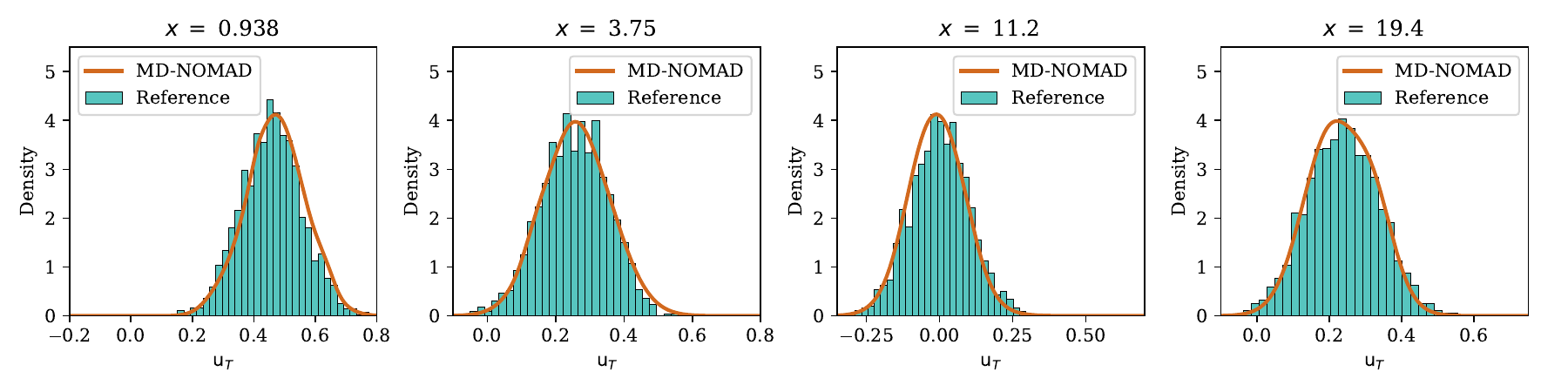}}
\caption{Comparison between reference distribution and MD-NOMAD predicted PDF at different spatial locations for \textbf{(a)} $\lambda = 0.519$ and \textbf{(b)} $\lambda = 0.741$ for 1-dimensional stochastic heat equation.}
\label{fig:12}
\end{figure}

\begin{figure}[ht!]
\centering
\subfigure[]{\label{subfig:lab131}\includegraphics[width=0.48\textwidth]{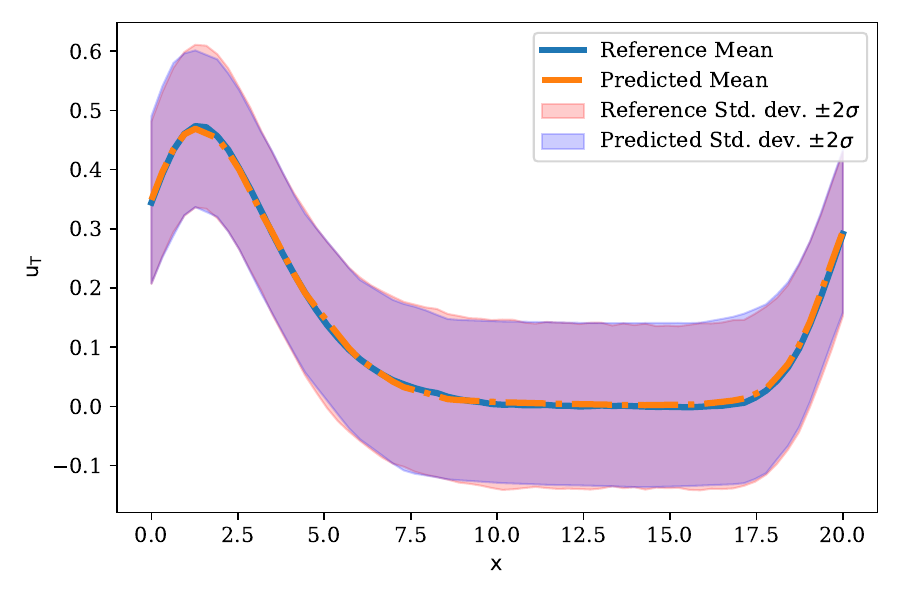}}
\subfigure[]{\label{subfig:lab132}\includegraphics[width=0.48\textwidth]{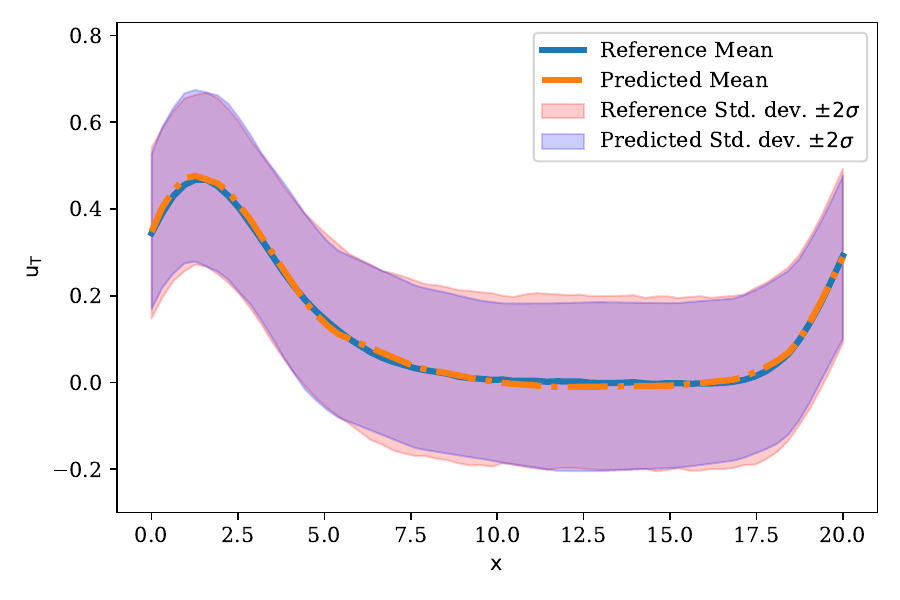}}
\caption{Comparison between reference statistics and MD-NOMAD predicted statistics for entire spatial domain $x\in[0,20]$ for \textbf{(a)} $\lambda = 0.519$ and \textbf{(b)} $\lambda = 0.741$ for 1-dimensional stochastic heat equation.}
\label{fig:13}
\end{figure}

The stochastic heat equation is a parabolic SPDE that provides a probabilistic framework for modeling the evolution of temperature distribution in a medium, considering both deterministic diffusive heat transfer and stochastic temperature fluctuations. For the current case, we consider a 1-dimensional stochastic heat equation, and the following equation represents it:

\begin{equation}
\begin{aligned}
   du(t, x) &= \alpha u_{xx}(t, x)dt + \lambda dW(t, x)\\
   \text{with } u(x,0) &= \left(1 + e^{-(2 - x)/\sqrt{2}}\right)^{-1} 
\end{aligned}  
\end{equation}

\noindent where \( u(t, x) \) represents the temperature at time \( t \) and position \( x \), \( \alpha > 0 \) is the thermal diffusivity, \( \lambda > 0 \) is the strength of the stochastic noise, and \( dW(t, x) \) denotes the Wiener process representing random fluctuations. The equation is solved using the pseudo-spectral method with Euler-Maruyama time integrator having a time step size \( \Delta t = 2.5\times10^{-3}\; \text{seconds} \) inside a spatial domain \( x \in [0,20] \) up to time $t=1$ second. Also, periodic boundary conditions are considered, and $\alpha=0.9$. The experimental design is established on the stochastic noise strength $\lambda$, which is sampled from the uniform distribution $\mathcal{U}(0.35,0.75)$, with $\mathscr{N}=50$ and $\mathscr{R} = 30$. Moreover, in this example, we are not interested in the time evolution of the state rather we are interested in the probability density for $u(x,t)$ at time $t=1$ s or $u_\mathrm{T}$. The branch network receives input $\lambda$, while the decoder network's additional input is the spatial location $x$. The training process consists of 300 epochs, utilizing 10 mixture components. Additionally, the target solution $u(x,1)$ is subjected to subsampling at a factor of 2, and corresponding input samples are accordingly preprocessed for the formation of training and validation datasets. The results that depict the performance of MD-NOMAD for the example are provided in Figure \ref{fig:12} and Figure \ref{fig:13}.

\subsection{1-dimensional stochastic viscous Burgers equation}
\begin{figure}[htbp!]
\centering
\subfigure[]{\label{subfig:lab141}\includegraphics[width=1.0\textwidth]{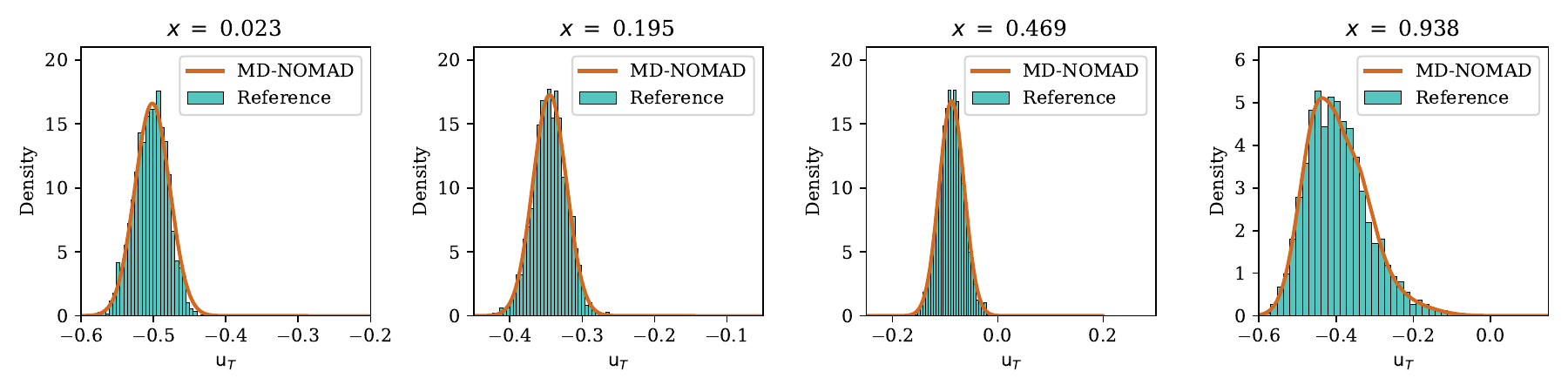}}
\subfigure[]{\label{subfig:lab142}\includegraphics[width=1.0\textwidth]{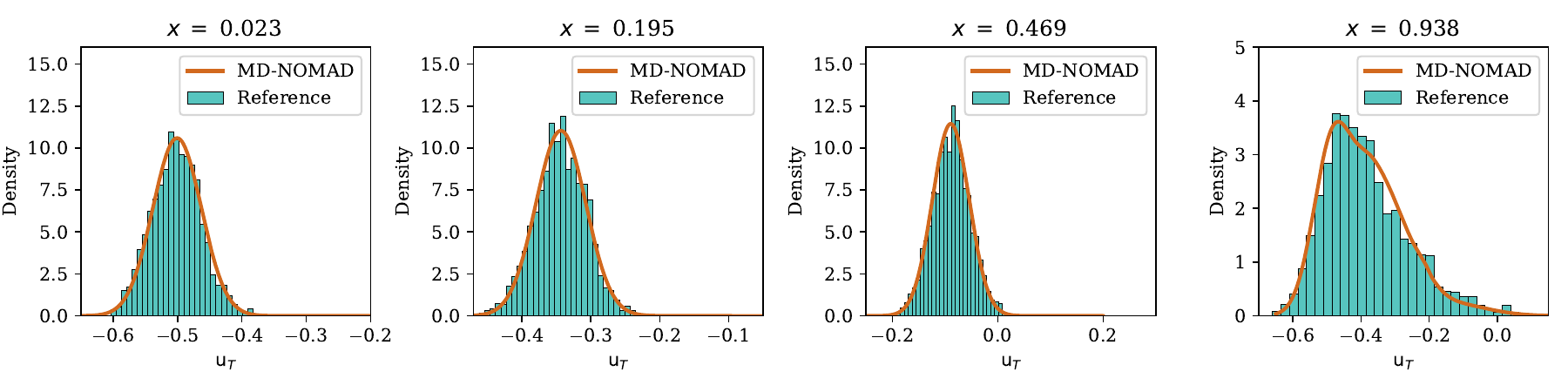}}
\caption{Comparison between reference distribution and MD-NOMAD predicted PDF at different spatial locations for \textbf{(a)} $\lambda = 0.452$ and \textbf{(b)} $\lambda = 0.72$ for 1-dimensional stochastic Burgers equation.}
\label{fig:14}
\end{figure}

\begin{figure}[htbp!]
\centering
\subfigure[]{\label{subfig:lab151}\includegraphics[width=0.48\textwidth]{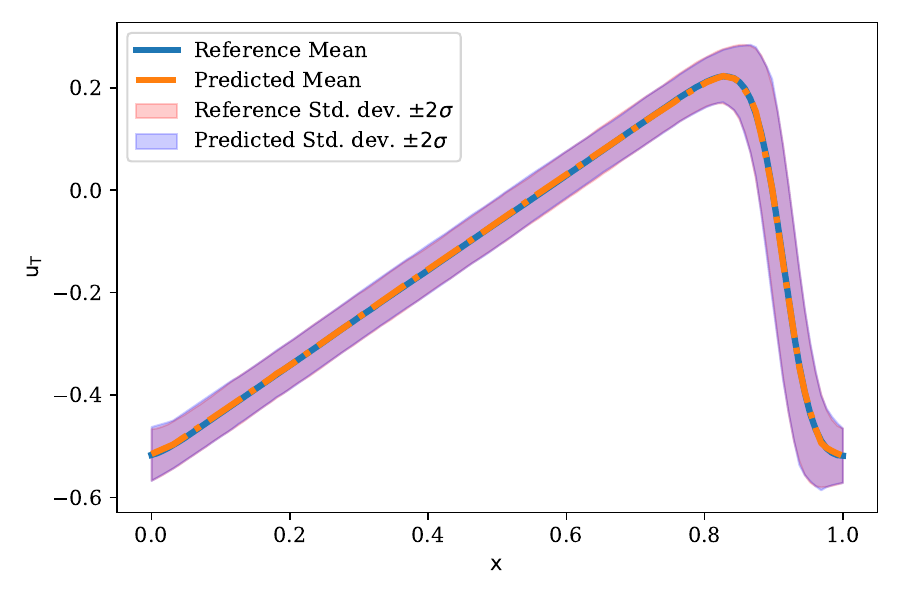}}
\subfigure[]{\label{subfig:lab152}\includegraphics[width=0.48\textwidth]{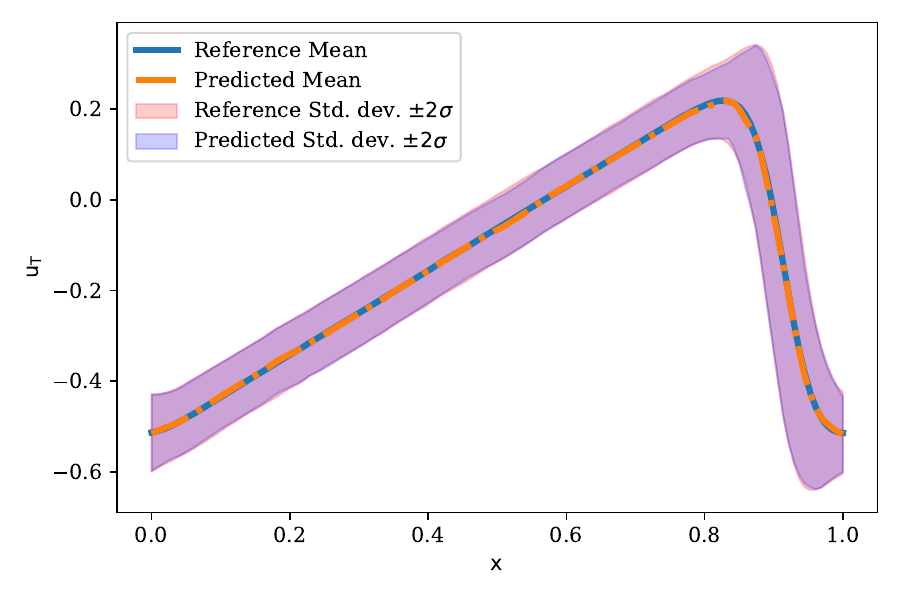}}
\caption{Comparison between reference statistics and MD-NOMAD predicted statistics for entire spatial domain $x\in[0,1]$ for \textbf{(a)} $\lambda = 0.452$ and \textbf{(b)} $\lambda = 0.72$ for 1-dimensional stochastic viscous Burgers equation.}
\label{fig:15}
\end{figure}

As our final example, we consider the 1-dimensional stochastic viscous Burgers' equation which is expressed as follows:
\begin{equation}
\begin{aligned}
   du(t, x) &= -u(t,x)\,u_{x}(t,x)dt + \mu\,u_{xx}(t, x)dt + \lambda\, dW(t, x),\\
   \text{with } u(0, x) &= \frac{2\, \phi(x)}{\max_{x}(\phi(x))} + \gamma,\; x\in[0,1],\; u(0,t) = u(1,t),\\
   \text{where }  \phi(x) &= a_0 + \sum_{k=1}^{2} a_k \sin(2 k\pi x) + b_k \cos(2 k\pi x),
\end{aligned}  
\end{equation}
where $a_k$, $b_k$, and $\gamma$ are randomly chosen fixed constants, \( u(t, x) \) represents the velocity at time \( t \) and position \( x \), \( \mu > 0 \) is the dynamic viscosity, \( \lambda > 0 \) is the stochastic noise strength, and \( dW(t, x) \) denotes the Wiener process. Again, the pseudo-spectral method is used along with the Euler-Maruyama time integrator for solving the SPDE. Further, the time step size \( \Delta t = 2.5\times10^{-4}\; \text{seconds} \) and $\mu = 0.01$. Moreover, even for this example, the experimental design is based on the stochastic noise strength $\lambda$, samples for which are drawn from the uniform distribution $\mathcal{U}(0.3,0.8)$, with $\mathscr{N}=50$ and $\mathscr{R} = 30$, and the quantity of interest is the solution's probability density at time $t=1$ s or $u_\mathrm{T}$. Also, the branch network takes in input $\lambda$, while the decoder network incorporates the spatial location $x$ as an additional input. The training process spans 300 epochs, incorporating 10 mixture components. Furthermore, the target solution $u(x,1)$ is exposed to subsampling with a factor of 2. Finally, Figure \ref{fig:14}  and Figure \ref{fig:15} depict the performance of MD-NOMAD for this example problem.

\begin{table}[ht!]
\caption{Comparison of expected Wasserstein distance of second-order (\(\mathcal{E}_W\)) and KL divergence (\(\mathcal{E}_{\text{KL}}\)) for MD-NOMAD and KCDE-based model across all example problems on an unseen test set.}
\centering
\begin{tabular}{l c c c c}
\hline\hline
\textbf{Problem} & \multicolumn{2}{c}{\textbf{MD-NOMAD}} & \multicolumn{2}{c}{\textbf{KCDE}} \\
 & \(\mathcal{E}_W\) & \(\mathcal{E}_{\text{KL}}\) & \(\mathcal{E}_W\) & \(\mathcal{E}_{\text{KL}}\) \\
\hline
Stochastic Van der Pol & $2.14 \times 10^{-1}$ & $7.33\times 10^{-2}$ & $6.09 \times 10^{-1}$ & $2.81 \times 10^{-1}$ \\
10-component stochastic Lorenz-96 & $7.26 \times 10^{-4}$ & $2.64\times 10^{-3}$ & $4.10 \times 10^{-3}$ & $1.05\times 10^{-1}$ \\
50-component stochastic Lorenz-96 & $6.28 \times 10^{-4}$ & $2.19 \times 10^{-3}$ & $3.76 \times 10^{-3}$ & $8.18\times 10^{-2}$ \\
Bimodal analytical benchmark & $1.55 \times 10^{-1}$ & $1.89\times 10^{-3}$ & $4.58 \times 10^{-1}$ & $2.74\times 10^{-2}$\\
2-d SPDE & $1.02 \times 10^{-4}$ & $8.12\times 10^{-3}$ & $1.24 \times 10^{-1}$ & $2.34$ \\
1-d stochastic heat & $5.46 \times 10^{-5}$ & $6.05\times 10^{-3}$ & $3.24 \times 10^{-4}$ & $5.45\times 10^{-2}$ \\
1-d stochastic viscous Burgers & $1.60 \times 10^{-5}$ & $4.23\times 10^{-3}$ & $8.88 \times 10^{-3}$ & $2.02$ \\
\hline\hline
\end{tabular}
\label{tab:model-comparison}
\end{table}

\section{Conclusions}
\label{S:6}
We introduced the mixture density-based NOMAD framework, which seamlessly integrates pointwise operator learning network NOMAD with mixture density formalism, enabling the estimation of conditional probability distributions for stochastic output functions. The components of the mixture were Gaussian distributions, which were predominantly employed for their mathematical traceability. The scalability inherited from the NOMAD architecture to handle high-dimensional problems, coupled with the ability to learn a data representation of stochastic systems, underscores the versatility and practical utility of MD-NOMAD. 
In our empirical assessment, we validated the efficacy of MD-NOMAD across a diverse range of stochastic ordinary and partial differential equations (including high-dimensional examples), covering both linear and nonlinear formulations, alongside analytical benchmark problems. In addition, MD-NOMAD was evaluated against the KCDE-based emulator, and as demonstrated in Table \ref{tab:model-comparison}, MD-NOMAD consistently outperforms KCDE across all benchmark problems. It is important to note that training a KCDE-based model becomes computationally demanding and increasingly complex for high-dimensional problems because of the curse of dimensionality. Furthermore, the usage of MD-NOMAD was also demonstrated for one-shot uncertainty propagation in a 2-dimensional stochastic partial differential equation. In addition, as the predicted PDFs and associated statistics for MD-NOMAD can be computed analytically, without resorting to MC simulations, the proposed framework experiences an improved computational efficiency, thereby bolstering its practical applicability. However, it is important to note that MD-NOMAD necessitates additional attention to adjusting an extra hyperparameter compared to deterministic NOMAD, specifically, the number of mixture components. Presently, the determination of the number of mixture components relies on validation error. However, Bayesian decision-making can be leveraged for automatic selection of the number of mixture components in later studies. Also, in this article, we employed a mixture of univariate Gaussian. While this choice offers several practical advantages, we acknowledge that it may limit the ability to accurately capture inter-component covariances. Investigating this limitation could be a subject of future work. Furthermore, future investigations into alternative architectures beyond the FCN utilized in this study, along with further endeavors such as assessing the capabilities of MD-NOMAD in constructing surrogates for complex 3-dimensional problems like turbulent channel flow, would be of considerable interest. Moreover, extending MD-NOMAD to function as a surrogate model in practical reliability analysis tasks --- such as for wind turbines, heritage structures, and composite plates --- could also be an interesting direction for future research.

\section*{Acknowledgement}
SC acknowledges the financial support received from the Ministry of Ports and Shipping via letter number ST-14011/74/MT (356529) and the Anusandhan National Research Foundation via grant number CRG/2023/007667.


\appendix
\renewcommand{\thesection}{Appendix \Alph{section}}
\section{Comparison between DDPM and MD-NOMAD for PDF inference effficiency} \label{app}
\appendix
In this appendix, we present a comparative analysis between MD-NOMAD and a DDPM, both trained to emulate a stochastic system characterized by the relation
\[
y = \sin(\pi x) + \epsilon,
\]
where \( x \sim \mathcal{U}(-1, 1) \) is a uniformly distributed input and \( \epsilon \sim \mathcal{N}(0, 0.1^2) \) denotes additive Gaussian noise. The goal is to learn the conditional PDF \( f_{Y|X}(y \mid x) \) and statistics, thereby enabling full probabilistic characterization of the system’s response at any arbitrary input location \( x \). Both MD-NOMAD and DDPM were trained on a dataset sampled from the stochastic process above. The two models differ significantly in their training and inference mechanisms. MD-NOMAD allows computation of the conditional PDF analytically by leveraging a parameterized mixture of distributions, thus enabling direct and efficient evaluation of \( f_{Y|X}(y \mid x) \) and corresponding statistics without the need for any sampling. This results in computational savings during PDF and statistic inference.

In contrast, obtaining the conditional PDF from the DDPM-based model requires generating a large number of samples for a given input condition via the learned reverse-time diffusion process, followed by kernel density estimation. This two-stage procedure makes PDF estimation/inference by the DDPM slower compared to the analytical approach used by MD-NOMAD. For this specific example, the wall clock time required to infer the conditional PDF at a given location \( x \) was found to be approximately 0.046 seconds for MD-NOMAD and 2.07 seconds for the DDPM-based model, indicating that MD-NOMAD achieves an approximately 4.5-fold increase in computational efficiency. Beyond its faster inference speed, MD-NOMAD also exhibits a lower hardware footprint, owing to the absence of complex sampling procedures and the elimination of the need to generate large numbers of samples. Furthermore, the error metric $\mathcal{E}_W$ was also evaluated for both models on a test set. It was found that MD-NOMAD achieved an error of $4.33 \times 10^{-5}$, in contrast to an error of $4.85 \times 10^{-4}$ observed for DDPM. This distinction might render MD-NOMAD particularly advantageous in real-time or resource-constrained environments and might also make it suitable for scenarios where rapid and scalable uncertainty quantification is critical.
% \begin{thebibliography}{10}
% \end{thebibliography}

\end{document}